\documentclass[runningheads]{llncs}
\usepackage[T1]{fontenc}
\usepackage{graphicx}
\usepackage{booktabs}
\usepackage[misc]{ifsym}
\newcommand{\corr}{(\Letter)}
\usepackage{adjustbox}
\usepackage{amsfonts,amsmath}
\usepackage{dsfont}
\usepackage{float}
\usepackage{hyperref}
\usepackage{lscape}
\usepackage{multirow}
\usepackage{tabularx}
\usepackage{tikz}

\newcolumntype{Y}{>{\centering\arraybackslash}X}

\newcommand*{\D}{\ensuremath{\mathcal{D}}}
\newcommand*{\Dt}{\ensuremath{\D_{\textnormal{train}}}}
\newcommand*{\Ds}{\ensuremath{\D_{\textnormal{synth}}}}
\newcommand*{\Dx}{\ensuremath{\D_{\textnormal{aux}}}}
\newcommand*{\Dg}{\ensuremath{\D_{\textnormal{target}}}}
\newcommand*{\Prob}{\ensuremath{\mathbb{P}}}
\newcommand*{\Phat}{\ensuremath{\mathbb{\hat{P}}}}
\newcommand{\Px}{\ensuremath{\Prob_X}}
\newcommand{\dX}{\mathcal{X}}
\newcommand*{\Dxs}{\ensuremath{\Dx^{\textnormal{ind}}}}
\newcommand*{\Dgs}{\ensuremath{\Dg^{\textnormal{ind}}}}
\newcommand*{\Dgh}{\ensuremath{\Dg^{\textnormal{house}}}}
\newcommand{\ind}[1]{\ensuremath{\mathds{1}\{#1\}}}

\begin{document}

\title{TAMIS: Tailored Membership Inference Attacks on Synthetic Data}
\author{
  Paul Andrey\inst{1} \corr
  \and Batiste Le Bars\inst{1}
  \and Marc Tommasi\inst{1}
}
\authorrunning{P. Andrey et al.}
\institute{
  Univ. Lille, Inria, CNRS, Centrale Lille,
  UMR 9189 - CRIStAL, F-59000 Lille, France
  \email{paul.andrey@inria.fr}
}

\maketitle

\begin{abstract}

Membership Inference Attacks (MIA) enable to empirically assess the privacy of
a machine learning algorithm.
In this paper, we propose TAMIS, a novel MIA against differentially-private
synthetic data generation methods that rely on graphical models.
This attack builds upon MAMA-MIA, a recently-published state-of-the-art method.
It lowers its computational cost and requires less attacker knowledge.
Our attack is the product of a two-fold improvement.
First, we recover the graphical model having generated a synthetic dataset
by using solely that dataset, rather than shadow-modeling over an auxiliary
one. This proves less costly and more performant.
Second, we introduce a more mathematically-grounded attack score, that provides
a natural threshold for binary predictions.
In our experiments, TAMIS achieves better or similar performance as MAMA-MIA on
replicas of the SNAKE challenge.


\end{abstract}

\newcommand\blfootnote[1]{%
  \begingroup
  \renewcommand\thefootnote{}\footnotetext{#1}%
  \endgroup
}
\blfootnote{
  \scriptsize
  Accepted at ECML PKDD 2025 (Research Track).\\
  First published in Lecture Notes in Computer Science (LNAI), volume 16017,
  pp 203-220, in 2025 by Springer Nature.
  Version of Record: \url{https://doi.org/10.1007/978-3-032-06096-9\_12}\\
  This expanded version includes appendices that detail experimental results.
}

\section{Introduction}

Synthetic Data Generation (SDG) consists in producing artificial samples that
match the specifications and retain distributional properties of actual data
from a given domain. Over the past decade, it has received increased focus as
a way to enable releasing data that can be used to learn statistics or even
train machine learning models without granting access to actual personal data.

However, research has shown that synthetic data is not inherently private.
Indeed, a synthetic dataset or a generative model learned from private
records can leak private information~\cite{mia:stadler}.
To mitigate this risk, most state-of-the-art SDG methods provide differential
privacy (DP) guarantees, that are formal properties of the method offering a
provable upper bound on the residual privacy risk~\cite{dp:dwork-roth}, usually
at the cost of decreased utility.
Most state-of-the-art SDG methods with DP guarantees rely either on learning a
graphical model to approximate the structure of the data
distribution~\cite{sdg:privbayes,sdg:mst,sdg:privmrf}, or on training
a generative neural network, typically
adversarially~\cite{sdg:dpgan,sdg:pategan}.

Privacy risks can also be assessed empirically using privacy attacks, which can
be complementary to DP guarantees~\cite{metrics:goodbadugly,metrics:anonymeter}.
Membership Inference Attack (MIAs) are a type of privacy attack where an
attacker having access to a trained machine learning model attempts to predict
whether certain records were part of its training data~\cite{mia:shokri2017}.
MIAs have been transposed to SDG by a number of
authors~\cite{mia:stadler,mia:hilprecht,mia:logan,mia:tapas,mia:domias}, who
introduced a variety of attacks and threat models.

The state-of-the-art method for MIA against SDG methods relying on graphical
models is MAMA-MIA~\cite{snake:mamamia}. It was developed to win the SNAKE
challenge~\cite{snake:paper}, where it achieved great success against the
MST~\cite{sdg:mst} and PrivBayes~\cite{sdg:privbayes} SDG methods, especially
in high-$\epsilon$ (that is, low-privacy) settings~\cite{snake:golob}.

In this paper, we introduce TAMIS, a novel attack that achieves better or
similar success as MAMA-MIA, has a lower computational cost and requires less
attacker knowledge.
We focus on attacking MST and PrivBayes, using principles that could be
extended to other graphical-model-based methods.
This new attack is the product of a two-fold improvement over MAMA-MIA.
On the one hand, we propose to recover the structure of the graphical model
having generated a synthetic dataset by using solely that dataset, rather than
shadow-modeling. This proves less costly and more performant, especially
against MST, for which we remove the need for the attacker to know any
hyper-parameter used for generation.
On the other hand, we introduce a more mathematically-grounded attack score,
that achieves similar or better performance as the MAMA-MIA one in experiments
over replicas of the SNAKE challenge. It notably achieves high accuracy without
requiring the attacker to know the true proportion of training points in the
attacked set.

In Section~\ref{sec:background} we provide more detailed background on
SDG and MIAs.
Then, we define the setting for our contributions in Section~\ref{sec:setting},
and come to present them in Section~\ref{sec:contribs}.
Experiments are reported in Sections~\ref{sec:expes} and~\ref{sec:results}.

\section{Background}
\label{sec:background}

\subsection{Synthetic Data Generation}

The aim of SDG is to produce a dataset of synthetic records $\Ds$ that follow a
similar distribution as observed private records in a training dataset $\Dt$.
In the remainder of this paper, we note $\Px$ the underlying distribution of
$\Dt$.

Numerous SDG methods have been proposed in the literature.
In this work, we focus on those based on graphical models as parametric
estimators of $\Px$, that approximated its dependency
structure~\cite{sdg:mst,sdg:privbayes,sdg:privmrf}.

Among other methods, we can mention those using generative neural networks as
non-parametric estimators of $\Px$~\cite{sdg:adsgan,sdg:dpgan,sdg:pategan}.
While less interpretable~\cite{study:auditability}, these methods cover a wider
variety of data types, including time series~\cite{sdg:doppelganger} or
multi-relational data~\cite{sdg:clavaddpm}.

Mechanisms have been introduced in SDG methods in order to provide DP
guarantees on the training data.
DP was introduced by Dwork \& Roth~\cite{dp:dwork-roth}, as a way to formalize
and quantify the privacy of an algorithm with respect to its input data.
Given $\epsilon > 0$ and $\delta \in ]0, 1[$, an algorithm
$\mathcal{A} : \D \rightarrow O$ is deemed $(\epsilon,\delta)$-differentially
private if, and only if, for any pair of adjacent datasets $\D, \D'$ (that is,
datasets that differ by a single record) and for any $S \subseteq O$,
$
  \Prob(\mathcal{A}(D) \in S)
  \leq
  e^{\epsilon} \Prob(\mathcal{A}(D') \in S) + \delta
$.

An important property of DP is the post-processing theorem.
For SDG, it means that the synthetic data inherit DP guarantees of
their generative model.

\subsection{Membership Inference Attacks}

MIAs were first defined by Shokri et al.~\cite{mia:shokri2017}.
In a classical MIA, the attacker has access to a trained model, and attempts
to predict whether certain records were part of its training dataset. In the
context of SDG, the attacker instead has access to a synthetic dataset $\Ds$,
knows some actual records $\Dg$ and tries to assess which of these were part
of $\Dt$ from which $\Ds$ was derived.

A variety of MIAs on SDG have been proposed in the
literature~\cite{mia:stadler,mia:hilprecht,mia:logan,mia:tapas,mia:domias},
that cover distinct threat models.
These were notably reviewed by Houssiau et al.~\cite{mia:tapas}, that
distinguish three main settings: in the white-box one, the attacker has full
access to the trained generative model; in the black-box one, they have
accurate knowledge of the SDG method; in the no-box one,
they only have access to a given $\Ds$.
Finer-grained variants of the black-box setting exist, notably as to
whether the attacker knows the hyper-parameters of the SDG method.

Another common hypothesis is for the attacker to have access to an auxiliary
dataset $\Dx$ that follows the same underlying distribution as $\Dt$.
With this, an attacker can notably perform shadow modeling, that
is run the SDG method on controlled inputs, resulting in labeled
replicas of the MIA setting~\cite{mia:stadler}.

\subsubsection{DOMIAS}~\cite{mia:domias}
is a generic framework to conduct MIAs on synthetic data, applicable to either
black-box or no-box settings but requiring access to auxiliary data.
DOMIAS aims to identify $\Dt$ samples that have been over-fitted by the SDG,
resulting in $\Ds$ concentrating more density around these samples
than a perfect estimate of $\Px$ would have.
Note that DP bounds and noises the contribution of samples,
hence offering protection against over-fitting and MIAs.

Attack scores are defined as the ratio of density functions estimated on
either $\Ds$ or $\Dx$, which we note $\Phat_X^{\Ds}$ and $\Phat_X^{\Dx}$:
\begin{equation}\label{eq:domias}
  \Lambda_\textnormal{DOMIAS}(x) = \cfrac{\Phat_X^{\Ds}(x)}{\Phat_X^{\Dx}(x)}
\end{equation}
This ratio is highest for samples that fit $\Ds$ more than $\Dx$, denoting
possible over-fitting.
Any density estimator may be plugged into this equation.

\subsubsection{MAMA-MIA}~\cite{snake:mamamia}
is explicitly inspired by DOMIAS, but requires black-box knowledge of the SDG
method and of all its hyper-parameters. It replaces density estimators with
statistics on $\Dt$ and $\Ds$ that are likely to have been selected by the
SDG method, hence computed on $\Dt$ and perpetuated in $\Ds$.
In other words, attack scores are made to focus on distributional features of
$\Px$ that were explicitly modeled by the SDG method.

To identify statistics on which to focus, MAMA-MIA uses shadow modeling of the
SDG method on $\Dx$, meaning it replicates the SDG method up to the statistics
selection step on random subsets of $\Dx$ that match the size of $\Dt$.

\section{Setting}
\label{sec:setting}

In this paper, we consider MIAs against SDG methods that rely on graphical
models. Our threat model is to assume access to a synthetic dataset $\Ds$ and
to an auxiliary dataset $\Dx \sim \Prob_X$, as well as black-box knowledge of
the nature of the SDG method, and when specified of its hyper-parameters.

We consider data with categorical attributes, hence a multivariate random
variable $X = (X_1, \dots, X_d)$
where $\forall i \in \{1, \dots, d\}, X_i \in \dX_i := \{1, \dots, n_i\}$
with $n_i \in \mathbb{N}^*$.
We note individual records in lower case: $x = (x_1, \dots, x_d)$.
Support for continuous variables can be achieved using quantization in pre-
and post-processing~\cite{sdg:privbayes,snake:mamamia}, which we leave
out without loss of generality.

A graphical model over $X$ is a family of probability distributions that can be
represented as a graph $G = (V, E)$, with nodes $V = \{1, \dots, d\}$ and
edges $E$ that define the structure of conditional dependencies between
attributes of $X$~\cite{graphs:wainwright-jordan}.
Given $G$, a specific distribution is obtained by defining some statistics,
based on which its joint density can be factorized.

The SDG methods we consider select a graphical model over $X$ that approximates
the structure of $\Px$, estimate associated statistics over $\Dt$, and generate
$\Ds$ as iid samples from the resulting distribution.
To achieve DP, randomness is added to both the graph selection and statistics
estimation steps. Figure~\ref{flow:sdg} summarizes this generic approach.

\usetikzlibrary{arrows, arrows.meta, chains, shapes}
\tikzstyle{stat} = [
  rectangle, draw, text width=10em, text centered,
  rounded corners, minimum height=3em
]
\tikzstyle{data} = [
  circle, draw, text width=3em, text centered, minimum height=3em
]
\begin{figure}[h]
  \caption{Flowchart of Synthetic Data Generation using a graphical model}\label{flow:sdg}
  \centering
  \begin{tikzpicture}[node distance=0.7cm and 1cm, start chain=going below]
    \node (dt) at (0,0) [data]  {$\Dt$};
    \node (gt) [stat, right=2.5cm of dt] {graph $G$};
    \node (pt) [stat, below=of gt] {distribution $\Phat_G^{\Dt}$};
    \node (ds) [data, right=2.5cm of pt] {$\Ds$};
    \draw [-{Latex[length=2mm]}] (dt) -- node[above] {select (with DP)} (gt);
    \draw [-{Latex[length=2mm]}] (gt) -- (pt);
    \draw [-{Latex[length=2mm]}] (dt) -- node[below left=-1mm and 0mm] {compute} node[below left=1mm and -6mm] {(noisy) statistics} (pt);
    \draw [-{Latex[length=2mm]}] (pt) -- node[above] {sample} (ds);
  \end{tikzpicture}
\end{figure}

\subsection{MST}

MST~\cite{sdg:mst} is a SDG method that relies on a tree graphical model.
A tree is a connected undirected graph with a constant number of edges
$|E| = |V| - 1$.
In such a graph, for any node $i$, we note $N(i) := \{j \in V | (i, j) \in E\}$
its neighbors.

The joint density of a tree graphical model is factorized based on the 1-way
marginals over its nodes and 2-way marginals over its edges, as
\begin{equation}
  \label{eq:density-tree}
  \Phat_G^{\D}(x) =
    \prod_{i \in V} \mu_i^{\D}(x)^{1 - |N(i)|}
    \prod_{(i,j) \in E} \mu_{ij}^{\D}(x)
\end{equation}
where $
\mu_i^{\D}(x)
    = \Phat_{\D}(X_i = x_i)
    := \frac{1}{|\D|} \sum_{\tilde{x} \in \D} \ind{\tilde{x}_i = x_i}
$
\\and
$
\mu_{ij}^{\D}(x)
  = \Phat_{\D}(X_i = x_i, X_j = x_j)
  := \frac{1}{|\D|} \sum_{\tilde{x} \in \D}
     \ind{\tilde{x}_i = x_i, \tilde{x}_j = x_j}
$.

\subsubsection{Graph selection in MST}
To select a tree graph from $\Dt$, MST uses a differentially-private
maximum-spanning tree algorithm. It assigns a score
$s_{ij} =
  \sum_{l_i, l_j \in \dX_i \times \dX_j}
    |
      \Phat_{\Dt}(X_i = l_i, X_j = l_j)
      - \Phat_{\Dt}(X_i = l_i)\Phat_{\Dt}(X_j = l_j)
    |
$ to each and every possible edge, with random noise added to 1-way marginals.
Then, at each of $|V|-1$ steps, an edge is randomly selected among valid
candidates with probabilities proportional to $s_{ij}$ scores.

\subsection{PrivBayes}

PrivBayes~\cite{sdg:privbayes} is a SDG method that relies on a bayesian
network.
A bayesian network is a directed acyclic graph. In such a graph, for any node
$i$, we note $\Pi_i := \{j \in V | (j, i) \in E\}$ its parent set, that is the
set of nodes with an edge towards $i$.
We note $x_{\Pi_i}$ the vector of coordinates of $x$ in $\Pi_i$.

The joint density of a bayesian network is factorized based on the conditionals
of nodes with respect to their parent set, as
\begin{equation}
  \label{eq:density-bnet}
  \Phat_G^{\D}(x) = \prod_{i \in V} \mu_{i,\Pi_i}^{\D}(x)
\end{equation}
where $
\mu_{i,\Pi_i}^{\D}(x)
  = \Phat_{\D}(X_i = x_i | X_{\Pi_i} = x_{\Pi_i})
  := \frac{1}{|\D|} \sum_{\tilde{x} \in \D} \ind{\tilde{x}_i = x_i, \tilde{x}_{\Pi_i} = x_{\Pi_i}}
$.

\subsubsection{Graph selection in PrivBayes}
To select a bayesian network from $\Dt$, PrivBayes uses a
differentially-private greedy algorithm.
At each of $|V|$ steps, a (node, parent set) tuple is randomly selected among
valid candidates with probabilities proportional to scores
$s_{i,\Pi_i} =
  \frac{1}{2}
  \sum_{l_i, \pi_i \in \dX_i \times \dX_{\Pi_i}}
    |
      \Phat_{\Dt}(X_i = l_i | \Pi_i = \pi_i)
      - \Phat_{\Dt}(X_i = l_i)\Phat_{\Dt}(\Pi_i = \pi_i)
    |
$.
Nodes and their parent set are constrained to have a total domain size below a
threshold that is proportional to the privacy budget $\epsilon$ and to a
hyper-parameter $\theta$, introducing a trade-off between structure-induced
approximation errors and DP-induced estimation errors.

\section{TAMIS: Tailored MIA on Synthetic data}
\label{sec:contribs}

In this paper, we introduce TAilored Membership Inference attacks on Synthetic
data (TAMIS), that are MIAs against SDG methods that rely on graphical models.
Our attack scores are based on the likelihood ratio approach of DOMIAS and
MAMA-MIA. The novelty of our approach is to learn the structure of a graphical
model matching the SDG method directly from $\Ds$, and to use the factorized
joint density of that model as a density estimator in attack scores.
Figure~\ref{flow:tamis} summarizes the TAMIS attack procedure as a flowchart.

\begin{figure}[h]
  \caption{Flowchart of the TAMIS attack}\label{flow:tamis}
  \centering
  \begin{tikzpicture}[node distance=0.5cm and 1cm, start chain=going below]
    \node (ds) at (0,0) [data] {$\Ds$};
    \node (gh) [stat, right=1.5cm of ds] {graph $\hat{G}$};
    \node (ps) [stat, below  left=0.7cm and -1.5cm of gh] {distribution $\Phat_G^{\Ds}$};
    \node (dx) [data, right=1.5cm of gh] {$\Dx$};
    \node (px) [stat, below right=0.7cm and -1.5cm of gh] {distribution $\Phat_G^{\Dx}$};
    \node (at) [stat, below right=0.7cm and -1.5cm of ps] {$\Lambda_{\textnormal{TAMIS}}$ score function};
    \node [above=1mm of at]{compose};
    \draw [-{Latex[length=2mm]}] (ds) -- node[above] {select} (gh);
    \draw [-{Latex[length=2mm]}] (ds) -- node[below left=-2mm and 0mm] {compute statistics} (ps);
    \draw [-{Latex[length=2mm]}] (gh) -- (ps);
    \draw [-{Latex[length=2mm]}] (dx) -- node[below right=-2mm and 0mm] {compute statistics} (px);
    \draw [-{Latex[length=2mm]}] (gh) -- (px);
    \draw [-{Latex[length=2mm]}] (px) -- (at);
    \draw [-{Latex[length=2mm]}] (ps) -- (at);
  \end{tikzpicture}
\end{figure}

By proposing an alternative to shadow modeling, we are able to lower the
computational cost of the attack compared with MAMA-MIA. For MST, we are also
able to remove the requirement for the attacker to know hyper-parameters.
Furthermore, by recovering a graph rather than a set of weights that cover
a broader set of edge choices, we are able to resort to a ratio of likelihoods
as attack score, which is distinct in nature from the MAMA-MIA attack scores,
as detailed below in this section.
Table~\ref{table:comparison} summarizes the key differences between the two
attacks.

\begin{table}[h]
  \centering
  \caption{Differences in hypotheses, costs and nature of TAMIS and MAMA-MIA}
  \label{table:comparison}
  \begin{tabularx}{\textwidth}{|cl|YY|YY|}
    \cline{1-6}
     & & \multicolumn{2}{c|}{MAMA-MIA}  & \multicolumn{2}{c|}{TAMIS} \\
     & & MST & PrivBayes & MST & PrivBayes \\
    \cline{1-6}
    \multirow[c]{4}{*}{\parbox{1.7cm}{\centering Graph\\Recovery}}
    & H: Known SDG params     & yes    & yes    & no       & yes \\
    & H: Known |$\Dt$|        & yes    & yes    & no       & no \\
    & H: Access to $\Dx$      & yes    & yes    & no       & no  \\
    & C: Cost relative to SDG & $K=50$ & $K=50$ & $\leq 1$ & 1   \\
    & N: Nature of the output & \multicolumn{2}{c|}{weights $W$} & \multicolumn{2}{c|}{graph $G$}\\
    \cline{1-6}
    \multirow[c]{2}{*}{\parbox{1.7cm}{\centering Attack\\Scores}}
    & H: Access to $\Dx$     & yes    & yes    & yes      & yes \\
    & N: Nature of the score & \multicolumn{2}{c|}{sum of statistics ratios} & \multicolumn{2}{c|}{ratio of densities}\\
    \cline{1-6}
  \end{tabularx}
\end{table}

\subsection{Graphical model recovery from the Synthetic Dataset}

The first step of our attack consists in selecting a graphical model that
matches that learned by the SDG method. Ideally, we would like to recover the
exact structure that was used for generating $\Ds$, in order to tailor attack
scores to features of $\Px$ that were actually (albeit noisily) measured on
$\Dt$.

To do so, we introduce SDG-method-specific algorithms that take $\Ds$ as input,
as opposed to the shadow-modeling approach of MAMA-MIA that is run on $\Dx$.
This decreases computational costs and avoids requiring access to $\Dx$.
We also believe this approach to be more rational. Indeed, shadow modeling on
$\Dx$ is bound to provide relatively generic information about the likely
structure of the generative model. On the opposite, that structure is bound to
be reflected in $\Ds$, hence easier to identify from it.
This point is especially important in DP regimes that introduce a lot of
randomness to the SDG graph selection step.

\subsubsection{Graph recovery for MST}

To recover the tree underlying $\Ds$ generated by MST, we introduce
a modified version of the graph selection step from MST, that is deprived of
DP mechanisms.
It consists in measuring all 1- and 2-way marginals of $\Ds$, assigning the
same edge-wise score as the MST selection algorithm does but without noise,
and finally using a maximum spanning tree algorithm to find the tree with the
highest possible sum of edge scores.

This algorithm is deterministic, involves slightly less computations than a
single shadow modeling run of the MST graph selection step, and does not
require any attacker knowledge. As such, it could be applied to any $\Ds$.

\subsubsection{Graph recovery for PrivBayes}

To recover the bayesian network underlying $\Ds$ generated by
PrivBayes, we simply apply the model-selection step of PrivBayes on $\Ds$.
This amounts to a single shadow modeling run, which is less costly than the
numerous ones run by MAMA-MIA (50 by default) but requires the same attacker
knowledge of the PrivBayes hyper-parameters, due to the selection step
adjusting the size of considered parent sets in the graph based
on these. It is also non-deterministic due to the DP mechanisms.

\subsection{Likelihood-based Attack Scores}

The second step of our attack consists in defining attack scores. To do so,
we use the graphical model resulting from the first step to estimate the
respective joint densities of $\Ds$ and $\Dx$, which requires computing some
statistics over both datasets. We then plug these densities into the DOMIAS
framework, meaning we use their ratio as attack scores.

Our scores are thus obtained by plugging the joint density formulas of
graphical models into the generic DOMIAS equation~\eqref{eq:domias}, that are
equation~\eqref{eq:density-tree} for MST and~\eqref{eq:density-bnet} for
PrivBayes. The resulting formulas are:

\begin{equation}\label{eq:domias-mst}
  \Lambda_{\textnormal{TAMIS-MST}}(x;G) =
    \prod_{i \in V}
      \left(\frac{\mu_i^{\Ds}(x)}{\mu_i^{\Dx}(x)}\right)^{1 - |N(i)|}
    \prod_{(i, j) \in E}
      \frac{\mu_{ij}^{\Ds}(x)}{\mu_{ij}^{\Dx}(x)}
\end{equation}
\begin{equation}\label{eq:domias-privbayes}
  \Lambda_{\textnormal{TAMIS-PB}}(x;G) =
    \prod_{i \in V} \frac{\mu_{i,\Pi_i}^{\Ds}(x)}{\mu_{i,\Pi_i}^{\Dx}(x)}
\end{equation}

TAMIS can therefore be thought of as an instantiation of the DOMIAS framework
that uses a graphical model as density estimator, and picks that graphical
model to mirror the one that was used in generating the synthetic data.
This follows the same intuition as MAMA-MIA, but enacts it in a different way.

\subsubsection{Comparison with MAMA-MIA}

In MAMA-MIA, the use of shadow modeling results in weights associated with
possible edge (for MST) or (node, parent set) (for PrivBayes) choices, rather
than in a valid graphical model.
After $K$ shadow runs, these weights are defined as
$\forall i \in \{1, \dots, d-1\}, \forall j \in \{i + 1, \dots, d\}$,
$w_{ij} = \sum_{k=1}^{K} \ind{(i, j) \in E^{(k)}}$ for MST, where
$E^{(k)}$ is the set of edges selected in the $k$-th shadow run.
For PrivBayes, they are defined as
$\forall i \in V, \forall \Pi_i \subset V\setminus\{i\}$,
$w_{i,\Pi_i} = \sum_{k=1}^{K} \ind{\forall j \in \Pi_i,\ (j, i) \in E^{(k)}}$.
We note $W$ a collection of such weights.

From there, the raw attack scores are defined as
\footnote{
  We note that the division of scores by $\sum_{w \in W} w$, amenable to
  normalizing weights, is an addition to the original MAMA-MIA formulas.
  We adopted it to avoid $K$ impacting the magnitude of raw scores.
  This improved predictions in our experiments.
}
\begin{equation}\label{eq:mamamia-mst}
  \Lambda_{\textnormal{MAMAMIA-MST}}(x;W) =
    \frac{1}{\sum_{w \in W} w}
    \sum_{w_{ij} \in W} w_{ij} \frac{\mu_{ij}^{\Ds}(x)}{\mu_{ij}^{\Dx}(x)}
\end{equation}
\begin{equation}\label{eq:mamamia-privbayes}
  \Lambda_{\textnormal{MAMAMIA-PB}}(x;W) =
    \frac{1}{\sum_{w \in W} w}
    \sum_{w_{i,\Pi_i} \in W} w_{i,\Pi_i}
      \frac{\mu_{i,\Pi_i}^{\Ds}(x)}{\mu_{i,\Pi_i}^{\Dx}(x)}
\end{equation}

We remark that these scores correspond to a weighted average of DOMIAS-like
scores attached to specific terms characterizing the density, whereas TAMIS
uses a DOMIAS-like score over the entire joint density.

We also note that when attacking MST, MAMA-MIA leaves apart information
from 1-way marginals, without justification. This omission is probably due to
the fact that 1-way marginals are always part of the graph, regardless of
selected edges.

\subsubsection{Hybrid scores}

We introduce hybrids of our graph-recovery approach with the formulas of
MAMA-MIA scores. This is useful to make more visible the difference between our
scores and the MAMA-MIA ones, and to disambiguate experimentally the impact
of the two folds of our contribution.
These scores come from replacing weights resulting from shadow modeling with
uniform weights that reflect the graph structure $G$ learned from $\Ds$
in Equations~\eqref{eq:mamamia-mst} and~\eqref{eq:mamamia-privbayes}.
\begin{equation}\label{eq:hybrid-mst}
\Lambda_{\textnormal{Hybrid-MST}}(x;G) =
  \frac{1}{|E|}
  \sum_{(i, j) \in E} \frac{\mu_{ij}^{\Ds}(x)}{\mu_{ij}^{\Dx}(x)}
\end{equation}
\begin{equation}\label{eq:hybrid-privbayes}
\Lambda_{\textnormal{Hybrid-PB}}(x;G) =
  \frac{1}{|V|}
  \sum_{i \in V} \frac{\mu_{i,\Pi_i}^{\Ds}(x)}{\mu_{i,\Pi_i}^{\Dx}(x)}
\end{equation}

\section{Experiments}
\label{sec:expes}

To assess the soundness and performance of our attacks, we conduct
experiments that replicate the SNAKE challenge \cite{snake:paper}.

We make our source code, data and random seeds available, enabling to fully
reproduce our experiments and results with minimal effort. They may be found
online\footnote{\url{https://gitlab.inria.fr/magnet/thesepaulandrey/tamis}},
together with dedicated documentation on implementation details.

In order to keep the paper short, we only detail main experimental results.
However, we make multiple references to appendices that provide with exhaustive
experimental results as well as some additional analyses.

\subsection{Dataset}

We used the publicly-available base dataset from the SNAKE challenge, that was
derived from US socio-demographic data published by the Economic Policy
Institute. It consists of about 201k samples with 15 variables, 12 of which are
categorical (with 2 to 51 possible labels each) and 3 of which have integer
values but can be treated as categorical nonetheless (age, number of children
and mean weekly number of worked hours). Additionally, individual samples each
belong to a given household, each of which groups 1 to 10 individuals.

We applied the same kind of preparation as was done in SNAKE. We take the base
dataset as $\Dx$, that is made available to the attacker. We randomly pick
100 households out of the 812 ones that group at least 5 individuals, and
designate their records as composing $\Dg$. Then, all individuals within half
of $\Dg$ households are made part of $\Dt$, and $\Dt$ is completed with random
individuals from $\Dx\setminus\Dg$ until $|\Dt| = 10 000$.

We generated 50 distinct replicas of the SNAKE $(\Dt, \Dg)$ generation.
We then ran both MST and PrivBayes on each and every replica with various
privacy budgets. For each setting, we generated $|\Ds| = |\Dt| = 10000$
synthetic samples, and recorded the structure of the graphical model that
generated them.
We considered $\epsilon \in \left[0.1, 1, 10, 100, 1000\right]$.
For MST, we always set $\delta = 10^{-9}$.
For PrivBayes, the DP mechanisms always achieve $\delta=0$.
For each value of $\epsilon$, we also ran shadow modeling of the graphical
model selection step of MST and PrivBayes on 50 random subsets of $\Dx$, as
was done in the MAMA-MIA paper.

\subsubsection{MIA evaluation settings}

We evaluate attacks on three distinct settings.
The first is $\Dxs$, where we attack each and every $\Dx$ sample, to assess the
average-case MIA risk.

The second is $\Dgh$, where we conduct set membership inference on households
of $\Dg$. This task is balanced by construction and matches original evaluation
setting of SNAKE. The raw score for a household is taken to be the average of
raw scores for samples in that household.
These households are expected to be easier to attack than any $\Dx$ sample.
Indeed, there is some correlation among samples in a household, hence when
they are jointly included in $\Dt$ they add weight to a given density region,
which can cause the SDG to over-fit that region.

The third is $\Dgs$, where we attack $\Dg$ samples. These are expected to be
somewhat easier targets than average samples for the reasons exposed before.

\subsection{Attack methods}

\subsubsection{Against MST} we compared three main attacks.
First, our TAMIS-MST attack~\eqref{eq:domias-mst}, using a tree graph learned
from $\Ds$.
Second, the MAMA-MIA attack targeted at MST~\eqref{eq:mamamia-mst}, using
weights resulting from shadow modeling over $\Dx$.
Finally, the Hybrid-MST attack~\eqref{eq:hybrid-mst}, that uses MAMA-MIA-like
scores over the same tree graph as TAMIS-MST.
Some additional baselines were considered to complement our comparison of
scores variants, that are reported in Appendix~\ref{appendix:mst-results}.

\subsubsection{Against PrivBayes} we compared three main attacks.
First, our TAMIS-PB attack~\eqref{eq:domias-privbayes}, using a bayesian
network learned from $\Ds$.
Second, the MAMA-MIA attack targeted at PrivBayes~\eqref{eq:mamamia-privbayes},
using weights resulting from shadow modeling over $\Dx$.
Finally, the Hybrid-PB attack~\eqref{eq:hybrid-privbayes}, that uses
MAMA-MIA-like scores over the same bayesian network as TAMIS-PB.

In addition, we introduced TAMIS-PB* and Hybrid-PB*, that use the same attack
scores as TAMIS-PB and Hybrid-PB respectively, but benefit from access to the
bayesian network that was truly selected by PrivBayes to generate $\Ds$. This
falls outside of our threat model, but is useful to assess the extent to which
recovering that graph can improve the success of MIAs.

\subsection{Evaluation}

\subsubsection{Graph Recovery}

We systematically compare the graphical model selected by the SDG methods to
generate $\Ds$ with that inferred from $\Ds$ as part of our attack. We compute
the accuracy of choices between the true and estimated graphs, comparing edges
for MST and (node, parent set) tuples for PrivBayes.

We also compare the edges or (node, parent set) tuples that are selected at
least once across shadow modeling runs, hence included in MAMA-MIA scores, with
true graphs. We compute precision, recall and jaccard index over both sets and
report their mean, standard deviation and median across replicas.

\subsubsection{Membership Inference Metrics}

To assess the success of MIAs, we report two metrics: the Area Under the
Receiver-Operator Curve (AUROC) of attack scores, and the balanced accuracy
of the resulting binary membership predictions.
The AUROC is invariant to the activation of raw attack scores, and enables
comparison with experimental results from the MAMA-MIA
paper~\cite{snake:mamamia}.
The balanced accuracy is defined as
$0.5 * (
    \frac{\Prob(\hat{Y} = 1 | Y = 1)}{\Prob(Y = 1)}  % true positive rate
  + \frac{\Prob(\hat{Y} = 0 | Y = 0)}{\Prob(Y = 0)}  % true negative rate
)$, where $Y = \ind{x \in \Dt}$ is the true membership label, and $\hat{Y}$ is
a binary prediction resulting from a raw attack score.
Mapping a raw attack score $\Lambda(x)$ into such a binary decision requires a
monotonous activation function $f$ that outputs predicted probabilities in
$[0, 1]$, and a threshold $t$ so that $\hat{Y} = \ind{f(\Lambda(x)) \geq t}$.

In practice, we consider two distinct activation and thresholding regimes.
In the \emph{simple} activation regime, we use a default threshold $t = 0.5$,
together with a classic activation function that is applied independently to
each score. We choose the sigmoid function, corrected to account for raw scores
being in $\mathbb{R}^+$: $f(x) = 2 * \textnormal{Sigmoid}(x) - 1$,
where $\textnormal{Sigmoid}(x) = (1 + e^{-x})^{-1}$.
In the \emph{calibrated} activation regime, we adopt the approach introduced
with MAMA-MIA, that was used to win the SNAKE challenge~\cite{snake:golob}.
We make the additional strong hypothesis that the attacker knows $\Prob(Y = 1)$
for the samples under attack, and adjusts the activation of these targets so that
$\Prob(\hat{Y} = 1) = \Prob(Y = 1)$.
First, raw attack scores are standardized into z-scores (that is, centered
around their mean then divided by their standard deviation), to avoid numerical
under- and over-flow issues. Then, z-scores are centered around their
$1 - \Prob(Y = 1)$ quantile, so that only a fraction $\Prob(Y = 1)$ are above
0. Finally, these scores are passed through the sigmoid function, and mapped
into decisions using $t = 0.5$.

\section{Results}
\label{sec:results}

\subsection{Attacks against MST}
\label{subsec:mst-results}

\begin{table}[t]
  \centering
  \caption{Main attack results against MST}
  \label{table:results-mst-short}
  \begin{tabular}{llccc}
  \toprule
   &  & \multicolumn{3}{c}{AUROC} \\
   &  & $\epsilon$=1000 & $\epsilon$=100 & $\epsilon$=10 \\
  \midrule
  \multirow[c]{3}{*}{\Dxs} & TAMIS-MST & \textbf{66.25}\ ($\pm$ 0.4) & \textbf{65.53}\ ($\pm$ 0.4) & 59.25\ ($\pm$ 0.4) \\
   & MAMAMIA-MST & 64.60\ ($\pm$ 1.0) & 64.13\ ($\pm$ 1.0) & 59.05\ ($\pm$ 0.6) \\
   & Hybrid-MST & 65.46\ ($\pm$ 0.3) & 64.97\ ($\pm$ 0.3) & \textbf{59.59}\ ($\pm$ 0.3) \\
  \cline{1-5}
  \multirow[c]{3}{*}{\Dgh} & TAMIS-MST & \textbf{77.76}\ ($\pm$ 4.8) & \textbf{77.44}\ ($\pm$ 3.6) & \textbf{69.87}\ ($\pm$ 5.0) \\
   & MAMAMIA-MST & 74.74\ ($\pm$ 5.4) & 74.62\ ($\pm$ 4.5) & 68.51\ ($\pm$ 5.4) \\
   & Hybrid-MST & 75.97\ ($\pm$ 4.6) & 75.57\ ($\pm$ 4.3) & 69.02\ ($\pm$ 5.5) \\
  \bottomrule
  \end{tabular}
  \begin{tabular}{llccc}
     &  & \multicolumn{3}{c}{Balanced Accuracy (Simple)} \\
   &  & $\epsilon$=1000 & $\epsilon$=100 & $\epsilon$=10 \\
  \midrule
  \multirow[c]{3}{*}{\Dxs} & TAMIS-MST & \textbf{60.84}\ ($\pm$ 0.3) & \textbf{60.39}\ ($\pm$ 0.3) & \textbf{55.66}\ ($\pm$ 0.3) \\
   & MAMAMIA-MST & 56.66\ ($\pm$ 0.7) & 56.44\ ($\pm$ 0.7) & 54.40\ ($\pm$ 0.4) \\
   & Hybrid-MST & 57.27\ ($\pm$ 0.4) & 57.06\ ($\pm$ 0.3) & 54.74\ ($\pm$ 0.3) \\
  \cline{1-5}
  \multirow[c]{3}{*}{\Dgh} & TAMIS-MST & \textbf{69.86}\ ($\pm$ 4.7) & \textbf{69.52}\ ($\pm$ 4.0) & \textbf{64.26}\ ($\pm$ 4.6) \\
   & MAMAMIA-MST & 61.06\ ($\pm$ 4.2) & 60.66\ ($\pm$ 3.4) & 58.16\ ($\pm$ 3.3) \\
   & Hybrid-MST & 61.76\ ($\pm$ 3.6) & 61.88\ ($\pm$ 3.7) & 58.08\ ($\pm$ 3.3) \\
  \bottomrule
  \end{tabular}
  \begin{tabular}{llccc}
     &  & \multicolumn{3}{c}{Balanced Accuracy (Calibrated)} \\
   &  & $\epsilon$=1000 & $\epsilon$=100 & $\epsilon$=10 \\
  \midrule
  \multirow[c]{3}{*}{\Dxs} & TAMIS-MST & \textbf{55.02}\ ($\pm$ 0.2) & \textbf{54.75}\ ($\pm$ 0.2) & 52.63\ ($\pm$ 0.2) \\
   & MAMAMIA-MST & 54.38\ ($\pm$ 0.6) & 54.15\ ($\pm$ 0.6) & 52.50\ ($\pm$ 0.4) \\
   & Hybrid-MST & 54.76\ ($\pm$ 0.2) & 54.54\ ($\pm$ 0.2) & \textbf{52.69}\ ($\pm$ 0.1) \\
  \cline{1-5}
  \multirow[c]{3}{*}{\Dgh} & TAMIS-MST & \textbf{70.24}\ ($\pm$ 5.2) & \textbf{69.88}\ ($\pm$ 3.4) & \textbf{64.26}\ ($\pm$ 4.7) \\
   & MAMAMIA-MST & 67.74\ ($\pm$ 4.8) & 67.24\ ($\pm$ 4.8) & 63.32\ ($\pm$ 5.1) \\
   & Hybrid-MST & 68.32\ ($\pm$ 4.8) & 68.06\ ($\pm$ 4.2) & 63.56\ ($\pm$ 5.4) \\
  \bottomrule
  \end{tabular}
\end{table}

\subsubsection{Graph recovery}

Our method accurately recovered the generating tree in all settings, achieving
a perfect match in edges selection for all replicas and $\epsilon$ value.
In comparison, as detailed in Appendix~\ref{appendix:graphs}, shadow modeling
missed a rarely-selected edge, and selected an increasing number of edges as
$\epsilon$ lowered due to DP, thus resulting in more un-tailored terms
in MAMA-MIA attack scores.

\subsubsection{Membership Inference}

Metrics of MST-targeted attacks on $\Dxs$ and $\Dgh$ are reported in
Table~\ref{table:results-mst-short}, excluding high-privacy regimes
$\epsilon \in [0.1, 1]$ for which no attack achieves significative success
due to the strong DP guarantees\footnote{
  The fact that attacks fail for lower $\epsilon$ values is expected
  given the theoretical guarantees. It is worth noting that these privacy
  guarantees come at a high utilty cost.
}.
Reported values are the average and standard deviation of metrics across
our 50 replicas.
We highlight the highest mean value per setting between attacks in bold.
Exhaustive results are placed in Appendix~\ref{appendix:mst-results}.

Results show that both of our contributions improve attack success.
First, Hybrid-MST achieves better results than MAMAMIA-MST, meaning that
replacing shadow modeling weights with the tree graph recovered from $\Ds$
improves MAMA-MIA-like attack scores.
Next, TAMIS-MST achieves even better success in nearly all settings, meaning
that the ratio of graphical model densities constitutes a better attack score
than the average of 2-way marginals ratios.

We remark that against $\Dxs$, TAMIS-MST and Hybrid-MST success metrics have a
markedly lower standard deviation than MAMAMIA-MST. We hypothesize that this is
due to shadow modeling weights causing MAMAMIA-MST scores to include terms that
variably match the actual generative graph of $\Ds$, and to exclude information
on a relevant edge for some replicas.
This validates the rationale of both TAMIS and MAMA-MIA to focus on
aspects of the distribution that were actively modeled during SDG.

Comparing balanced accuracy across activation regimes, we observe on $\Dgh$
that TAMIS-MST, which is more accurate than others in both settings, receives
less improvement from calibration. Hence TAMIS-MST attack scores appear to be
naturally suitable for sigmoid activation with a basic threshold. Oppositely,
MAMA-MIA-like scores appear to rely on calibration, hence on an additional
piece of attacker knowledge, to be turned into accurate predictions against
$\Dgh$.
We also observe that calibration on the unbalanced $\Dxs$ results in a
decrease in balanced accuracy for all attack scores. This hints that the
calibration proposed by MAMA-MIA authors may be over-fitted to the balanced
$\Dgh$ setting, that was the target of the SNAKE competition.

\subsection{Attacks against PrivBayes}
\label{subsec:privbayes-results}

\subsubsection{Graph recovery}

Our method almost never perfectly recovered the generating bayesian network
from $\Ds$. As detailed in Appendix~\ref{appendix:graphs}, for higher
$\epsilon$ values, about half (node, parentset) choices match, while mismatches
arise from marginal differences in including this or that edge, which can have
a strong impact on the modeled density.
In comparison, shadow modeling achieves above 90~\% recall, but less than 25~\%
precision for $\epsilon \geq 1$, meaning MAMA-MIA attack scores contain most
conditionals attached to the generative density of $\Ds$, but at least 3 times
more terms made of other conditionals.
Interestingly, graph recovery is easier when $\epsilon$ lowers, due to
PrivBayes reducing possible choices.

\subsubsection{Membership Inference}

Metrics of PrivBayes-targeted attacks on $\Dxs$ and $\Dgh$ are reported in
Table~\ref{table:results-privbayes-short}, excluding high-privacy regimes
$\epsilon \in [0.1, 1]$ for which no attack achieves significative success
due to the strong DP guarantees. We also excluded balanced accuracy under the
calibrated activation regime, which is of lesser interest as rapidly discussed
below.
Exhaustive results are placed in Appendix~\ref{appendix:privbayes-results}.
Reported values are the average and standard deviation of metrics across our
50 replicas.
For each setting, we highlight two values in bold, that are the highest
mean value either between TAMIS-PB, MAMAMIA-PB and Hybrid-PB or between
TAMIS-PB* and Hybrid-PB*.

\begin{table}[t]
  \centering
  \caption{Main attack results against PrivBayes}
  \label{table:results-privbayes-short}
  \begin{tabular}{llccc}
  \toprule
   &  & \multicolumn{3}{c}{AUROC} \\
   &  & $\epsilon$=1000 & $\epsilon$=100 & $\epsilon$=10 \\
  \midrule
  \multirow[c]{5}{*}{\Dxs} & TAMIS-PB & 64.65\ ($\pm$ 2.1) & 62.65\ ($\pm$ 1.3) & 53.99\ ($\pm$ 0.3) \\
   & MAMAMIA-PB & \textbf{79.34}\ ($\pm$ 1.5) & 64.36\ ($\pm$ 1.1) & \textbf{54.47}\ ($\pm$ 0.3) \\
   & Hybrid-PB & 79.33\ ($\pm$ 2.7) & \textbf{64.48}\ ($\pm$ 1.7) & 54.42\ ($\pm$ 0.4) \\
   \cline{2-5}
   & TAMIS-PB* & 66.72\ ($\pm$ 0.8) & 64.61\ ($\pm$ 0.4) & 54.16\ ($\pm$ 0.3) \\
   & Hybrid-PB* & \textbf{83.01}\ ($\pm$ 0.9) & \textbf{67.06}\ ($\pm$ 0.3) & \textbf{54.72}\ ($\pm$ 0.2) \\
  \cline{1-5}
  \multirow[c]{5}{*}{\Dgh} & TAMIS-PB & 82.74\ ($\pm$ 5.6) & 76.85\ ($\pm$ 5.2) & \textbf{62.00}\ ($\pm$ 5.4) \\
   & MAMAMIA-PB & \textbf{92.16}\ ($\pm$ 2.6) & 78.06\ ($\pm$ 5.5) & 61.53\ ($\pm$ 4.7) \\
   & Hybrid-PB & 91.87\ ($\pm$ 3.0) & \textbf{78.37}\ ($\pm$ 5.6) & 61.49\ ($\pm$ 5.5) \\
   \cline{2-5}
    & TAMIS-PB* & 85.75\ ($\pm$ 4.2) & 80.47\ ($\pm$ 4.8) & \textbf{62.64}\ ($\pm$ 4.7) \\
   & Hybrid-PB* & \textbf{94.32}\ ($\pm$ 2.6) & \textbf{81.98}\ ($\pm$ 4.8) & 62.23\ ($\pm$ 4.9) \\
  \bottomrule
  \end{tabular}
  \begin{tabular}{llccc}
     &  & \multicolumn{3}{c}{Balanced Accuracy (Simple)} \\
   &  & $\epsilon$=1000 & $\epsilon$=100 & $\epsilon$=10 \\
  \midrule
  \multirow[c]{5}{*}{\Dxs} & TAMIS-PB & 61.67\ ($\pm$ 1.8) & \textbf{58.53}\ ($\pm$ 1.1) & \textbf{52.48}\ ($\pm$ 0.3) \\
   & MAMAMIA-PB & 69.97\ ($\pm$ 1.4) & 55.49\ ($\pm$ 0.6) & 50.81\ ($\pm$ 0.1) \\
   & Hybrid-PB & \textbf{70.75}\ ($\pm$ 2.4) & 56.21\ ($\pm$ 0.9) & 50.85\ ($\pm$ 0.1) \\
   \cline{2-5}
   & TAMIS-PB* & 63.38\ ($\pm$ 0.7) & \textbf{60.18}\ ($\pm$ 0.3) & \textbf{52.68}\ ($\pm$ 0.2) \\
   & Hybrid-PB* & \textbf{74.24}\ ($\pm$ 1.0) & 57.67\ ($\pm$ 0.4) & 50.90\ ($\pm$ 0.1) \\
  \cline{1-5}
  \multirow[c]{5}{*}{\Dgh} & TAMIS-PB & 76.04\ ($\pm$ 5.5) & \textbf{70.20}\ ($\pm$ 4.8) & \textbf{57.04}\ ($\pm$ 4.2) \\
   & MAMAMIA-PB & 78.54\ ($\pm$ 4.2) & 57.34\ ($\pm$ 3.7) & 51.74\ ($\pm$ 2.5) \\
   & Hybrid-PB & \textbf{82.14}\ ($\pm$ 4.9) & 59.86\ ($\pm$ 3.6) & 51.70\ ($\pm$ 2.4) \\
   \cline{2-5}
   & TAMIS-PB* & 79.48\ ($\pm$ 4.8) & \textbf{73.12}\ ($\pm$ 4.5) & \textbf{57.64}\ ($\pm$ 3.5) \\
   & Hybrid-PB* & \textbf{86.58}\ ($\pm$ 4.0) & 62.36\ ($\pm$ 3.4) & 51.54\ ($\pm$ 2.3) \\
  \bottomrule
  \end{tabular}
\end{table}

We first compare attacks that are in line with our threat model, namely
TAMIS-PB, Hybrid-PB and MAMAMIA-PB.
Results are somewhat ambivalent.
On the one hand, when considering the balanced accuracy with simple activation
on either $\Dxs$ or $\Dgh$, TAMIS-PB achieves the best results of all methods,
save for the lowest-privacy regime $\epsilon=1000$, where it is the worst.
On the other hand, when considering the AUROC, TAMIS-PB is most often the worst
method. On $\Dxs$, MAMAMIA-PB achieves the best results, followed by Hybrid-PB
that has similar average values but higher variance.
On $\Dgh$, Hybrid-PB and MAMAMIA-PB have similarly good results, while TAMIS-PB
is worse or equal, with lower differences towards the other methods than
on $\Dxs$.

Next, we compare the TAMIS-PB* and Hybrid-PB* attacks, that are granted
knowledge of the true bayesian network. We retrieve the same ambivalence as to
which attack score is best: Hybrid-PB* has a higher or similar AUROC as
TAMIS-PB*, while it has a markedly lower balanced accuracy using simple
activation, save for the $\epsilon=1000$ where it is markedly better.
We also observe that TAMIS-PB* and Hybrid-PB* achieve better performance than
the other three attacks. This again validates the hypothesis that focusing on
distributional features selected by the SDG is key in crafting efficient attack
scores. While it is unlikely that an attacker would be directly provided
knowledge of the bayesian network structure that generated $\Ds$, our results
highlight that improving over our proposed graph recovery method could enable
out-performing both the MAMA-MIA and current TAMIS-PB attacks.

Regarding calibrated activation, we renew the observations made for MST.
On the one hand, this activation is detrimental to the accuracy of predictions
on $\Dxs$ for all methods. On the other hand, on the specific $\Dgh$ setting,
it is markedly beneficial to the accuracy of predictions for MAMA-MIA-like
scores (MAMAMIA-PB and Hybrid-PB) while having very limited impact on TAMIS-PB,
that still achieves better or similar balanced accuracy in most cases.

Factoring all previous results, we conclude that while MAMA-MIA-like scores
seemingly extract more information than TAMIS-PB, as showed by their higher
AUROC, the TAMIS-PB scores are more suitable for prediction in the realistic
setting where the attacker has no additional knowledge to craft decision
thresholds beyond the default $t = 0.5$.

We hypothesize that TAMIS scores underperforming for $\epsilon=1000$
may be due to numerical effects of extreme values in ratios of
conditionals. Notably, when attacking a sample that exhibits a combination
of attributes unseen in either $\Dx$ or $\Ds$, both the MAMA-MIA and TAMIS
implementations assign an arbitrarily-low conditional probability to it, which
is bound to result in extreme ratio values. This is more likely to appear with
high $\epsilon$, as parent sets are authorized to have a larger domain size,
possibly containing very-rare combinations. This could probably be addressed
by refining the way these cases are handled.

\subsection{Cross-targeted attacks}
\label{subsec:cross-results}

In order to further assess how beneficial it is for attacks to be tailored to
the SDG method, we experimented running MST-targeted attacks against
PrivBayes-generated synthetic data, and conversely.
Results are provided in Appendix~\ref{appendix:crosstarget-results}.

Overall, cross-targeted attacks under-perform compared with their counterparts.
Notably, PrivBayes-targeted attacks have poor results against MST-generated
data. This highlights that using a more complex density estimator does not
necessarily result in more performant attacks, again validating the value
of using a density estimator matching the generating one.

However, TAMIS-MST achieves good results against PrivBayes, with a higher or
similar balanced accuracy with simple activation than MAMAMIA-PB and Hybrid-PB
for all settings, save for $\epsilon=1000$.
This is remarkable, as TAMIS-MST requires less attacker knowledge and
computational power than MAMA-MIA.

\section{Conclusion}

In this paper, we have investigated and validated the assumption that tailoring
MIAs on synthetic data to the SDG method was beneficial to attack performance,
and proposed alternatives to refine both steps of the state-of-the-art MAMA-MIA
attack. The resulting TAMIS attacks were demonstrated to further improve the
state-of-the-art against MST and PrivBayes on replicas of the SNAKE challenge,
that MAMA-MIA recently won.

Our experiments have shown that recovering the graphical model underlying a
synthetic dataset resulted in more successful attacks than gathering more
generic information via shadow modeling. For MST, we were able to propose
a straightforward graph recovery method that achieved perfect accuracy in
our experiments, and requires both less computational power and less attacker
knowledge than shadow modeling. For PrivBayes, our method only improves over
shadow modeling in terms of computational power, and should be a focus for
improvement in future work.

Our experiments have also shown that our proposed attack scores, which are
more mathematically-grounded than their MAMA-MIA counterpart, produce more
accurate predictions when using a simple sigmoid activation function and a
default decision threshold. We have also shown that methods are not ranked
similarly depending on whether their AUROC or balanced accuracy is considered.
This highlights that there may be a gap between the information contained in
attack scores and that which can instrumented into actual predictions by an
attacker. Future research may help close that gap, either by designing clever
activation and thresholding mechanisms that do not rely on unrealistic attacker
knowledge hypotheses, or by further refining the way how focused statistics are
combined into attack scores that both achieve high AUROC and behave nicely with
simple activation functions.

Finally, while the MAMA-MIA and TAMIS attacks have been designed for the
black-box model threat where the attacker knows which SDG method was used,
we have remarked that the TAMIS-MST attack could in fact be run agnostic to
the SDG method. Our experiments on PrivBayes-generated data have shown that
it may be a competitive if not optimal method, and a future research direction
could be to assess the value of that attack in a no-box threat model, using
more diverse datasets, SDG methods and relevant baseline attacks.

\begin{credits}
\subsubsection{\ackname}
This work was supported by the
ANR 22-CMAS-0009 CAPS'UL (CAmpus Participatif en Santé numérique du Site
Universitaire de Lille) project of the France 2030 AMI-CMA,
and the ANR 22-PECY-0002 IPOP (Interdisciplinary Project on Privacy) project
of the Cybersecurity PEPR.

\subsubsection{\discintname}
The authors have no competing interests to declare that are relevant to the
content of this article.
\end{credits}

\bibliographystyle{splncs04}
\bibliography{bibliography}

\clearpage
\appendix

\section{Graph Recovery Success}
\label{appendix:graphs}

In this appendix, we present evaluation metrics on the adequation between
graphs selected either by running graph-selection algorithms on $\Ds$ or
shadow modeling on $\Dx$, and the actual graph selected by the SDG method
to generate $\Ds$.

Table~\ref{table:graph-mst} presents results for our proposed MST graph
selection algorithm, that takes $\Ds$ as input, without any knowledge of SDG
hyper-parameters. It shows that the method systematically achieved perfect
recovery of the graph structure for MST: the list of selected edges matches
exactly that of the graph that generated $\Ds$.

\begin{table}
  \centering
  \caption{MST graph recovery success metrics}
  \label{table:graph-mst}
  \begin{tabular}{lccc}
  \toprule
   & \multicolumn{2}{c}{Accuracy} & Perfect matches \\
   & mean & stdv &  \\
  \midrule
  $\epsilon=1000$ & 100.00 & 0.00 & 100.00 \\
  $\epsilon=100$ & 100.00 & 0.00 & 100.00 \\
  $\epsilon=10$ & 100.00 & 0.00 & 100.00 \\
  $\epsilon=1$ & 100.00 & 0.00 & 100.00 \\
  $\epsilon=0.1$ & 100.00 & 0.00 & 100.00 \\
  \bottomrule
  \end{tabular}
\end{table}

In comparison,
table~\ref{table:sm-graph-mst} presents adequation metrics between the set of
edges that were selected at least once by MST shadow modeling on $\Dx$ random
subsets with the edges of actual generative graphs for $\Ds$ for the same
privacy budget.
We report precision, defined as $P((i,j) \in E | (i,j) \in \tilde{E})$; recall,
defined as $P((i,j) \in \tilde{E} | (i,j) \in E)$; and jaccard index, defined
as $|\tilde{E} \cap E| / |\tilde{E} \cup E|$.
In these definitions, $E$ is the edge list of the true generative graph for
a given $\Ds$ while $\tilde{E}$ is the set of edges selected across shadow runs.

Results show that shadow modeling selects an increasing number of
distinct edges when $\epsilon$ is lowered, which is coherent with DP inducing
more randomness in the selection process to obtain stronger privacy guarantees.
This is reflected by the decreasing precision and jaccard index, all the while
recall remains stable.
As a consequence, MAMA-MIA attack scores factor information that is broader,
hence less tailored to the actual $\Ds$ under attack, when $\epsilon$ is lower.
We note that for all $\epsilon$ choices, recall is very close but not equal to
100~\% on average, which indicates that at least one edge that is selected by
the SDG method for one or few replicas was never selected across the shadow
modeling runs.

\begin{table}
  \centering
  \caption{MST shadow modeling graph recovery metrics}
  \label{table:sm-graph-mst}
  \begin{tabular}{l|ccc|ccc|ccc}
  \toprule
   & \multicolumn{3}{c|}{precision} & \multicolumn{3}{c|}{recall} & \multicolumn{3}{c}{jaccard} \\
   & mean & stdv & median & mean & stdv & median & mean & stdv & median \\
  \midrule
  $\epsilon=1000$ & 77.56 & 1.10 & 77.78 & 99.71 & 1.41 & 100.00 & 77.40 & 1.85 & 77.78 \\
  $\epsilon=100$ & 81.88 & 1.61 & 82.35 & 99.43 & 1.96 & 100.00 & 81.54 & 2.78 & 82.35 \\
  $\epsilon=10$ & 77.56 & 1.10 & 77.78 & 99.71 & 1.41 & 100.00 & 77.40 & 1.85 & 77.78 \\
  $\epsilon=1$ & 57.67 & 1.54 & 58.33 & 98.86 & 2.65 & 100.00 & 57.32 & 2.35 & 58.33 \\
  $\epsilon=0.1$ & 35.74 & 0.62 & 35.90 & 99.57 & 1.71 & 100.00 & 35.69 & 0.82 & 35.90 \\
  \bottomrule
  \end{tabular}
\end{table}

Table~\ref{table:graph-privbayes} presents results for our proposed PrivBayes
graph selection algorithm, that takes $\Ds$ as input and requires knowledge
of the hyper-parameters used to generate it. The method amounts to a single
shadow modeling run on $\Ds$.

Results show that for $\epsilon \geq 1$, the graph selected from $\Ds$ nearly
never fully matches the generative one. This is due to the complexity of
selection in PrivBayes, that picks (node, parent set) tuples defining complex
conditionals, and not simple edges as MST. Further computations show that the
precision and recall associated with edge choices are in fact around 65~\%
on average, with little variation across $\epsilon$ values, hinting that the
selected parent sets are not incoherent between the true and predicted graphs
even when they are not equal.

The accuracy of choices lies between 40 and 60~\% for $\epsilon \geq 10$,
meaning that about half (node, parent set) choices in predicted graphs match
the generative one. Thus, TAMIS-PB and Hybrid-PB attacks factor conditionals
that are not perfectly tailored to the attacked dataset, with about half the
score terms relying on partially-distinct parent sets.

We see that that the accuracy of choices increases when $\epsilon$ lowers,
reaching nearly 70~\% and more than 90~\% for $\epsilon=1$ and $\epsilon=0.1$
respectively. We alo see that for $\epsilon=0.1$ the graph is perfectly
recovered in 23 out of 50 replicas.
This may seem surprising as a lower $\epsilon$ is supposed to induce more
privacy, but results from the DP budget reducing the amount of valid (node,
parent set) tuples in the PrivBayes graph selection algorithm, that restricts
their total domain size based on $\epsilon$. With fewer possible edges to
choose from, the probability that choices match is mechanically higher.

\begin{table}
  \centering
  \caption{PrivBayes graph recovery success metrics}
  \label{table:graph-privbayes}
  \begin{tabular}{lccc}
  \toprule
   & \multicolumn{2}{c}{Accuracy} & Perfect matches \\
   & mean & stdv &  \\
  \midrule
  $\epsilon=1000$ & 43.60 & 21.26 & 2.00 \\
  $\epsilon=100$ & 41.60 & 19.45 & 0.00 \\
  $\epsilon=10$ & 55.87 & 18.75 & 0.00 \\
  $\epsilon=1$ & 68.27 & 11.47 & 2.00 \\
  $\epsilon=0.1$ & 92.80 & 6.71 & 46.00 \\
  \bottomrule
  \end{tabular}
\end{table}

Table~\ref{table:sm-graph-privbayes} presents adequation metrics between the
set of (node, parentset) tuples that were selected at least once by PrivBayes
shadow modeling on $\Dx$ random subsets with those defining the actual
generative bayesian network for $\Ds$ for the same privacy budget.
The metrics are similar to those presented above for MST, but compare
(node, parentset) tuple choices rather than edge choices.
We observe that adequation metrics improve when $\epsilon$ is lower, which is
due to the way PrivBayes restricts possible choices based on the DP budget, as
we previously noted.

The recall is above 90~\% on average for $\epsilon=1000$, and reaches 100~\%
for $\epsilon=0.1$, hence the conditionals that compose the MAMA-MIA attack
scores nearly always contain those that define the generating density of $\Ds$.
Precision, on the other hand, is rather low, save for $\epsilon=0.1$. This
indicates that a broader set of choices is made with higher $\epsilon$, again
due to the combinatorial effect, combined with DP-induced randomness, that can
play a role even for high $\epsilon$ when multiple solutions have similar
information scores. Hence, we expect MAMA-MIA scores to be all the more
tailored to $\Ds$ when $\epsilon$ is low; but in these cases, the DP-induced
noise on conditionals makes it nearly impossible that they reflect over-fitting
of $\Dt$ samples.

\begin{table}
  \centering
  \caption{PrivBayes shadow modeling graph recovery metrics}
  \label{table:sm-graph-privbayes}
  \begin{tabular}{l|ccc|ccc|ccc}
  \toprule
   & \multicolumn{3}{c|}{precision} & \multicolumn{3}{c|}{recall} & \multicolumn{3}{c}{jaccard} \\
   & mean & stdv & median & mean & stdv & median & mean & stdv & median \\
  \midrule
  $\epsilon=1000$ & 10.78 & 1.50 & 11.51 & 90.53 & 12.57 & 96.67 & 10.68 & 1.61 & 11.46 \\
  $\epsilon=100$ & 13.64 & 0.57 & 14.02 & 97.33 & 4.04 & 100.00 & 13.60 & 0.64 & 14.02 \\
  $\epsilon=10$ & 22.33 & 0.80 & 22.73 & 98.27 & 3.51 & 100.00 & 22.25 & 0.96 & 22.73 \\
  $\epsilon=1$ & 24.50 & 0.77 & 25.00 & 98.00 & 3.09 & 100.00 & 24.39 & 0.95 & 25.00 \\
  $\epsilon=0.1$ & 88.24 & 0.00 & 88.24 & 100.00 & 0.00 & 100.00 & 88.24 & 0.00 & 88.24 \\
  \bottomrule
  \end{tabular}
\end{table}

\clearpage

\section{Results of attacks against MST}
\label{appendix:mst-results}

This appendix presents exhaustive experimental results about the success of
MST-targeted attacks, that are reported in table~\ref{table:results-mst-full}.
Reported values are the average and standard deviation of metrics across
50 replicas.
We highlight the highest mean value per setting between attacks in bold.

The evaluated attacks comprise those presented in the main paper: TAMIS-MST,
Hybrid-MST and MAMAMIA-MST. The first two rely on a tree learned by the
attacker from $\Dt$, that is identical to the actual generative tree. The third
relies on 50 shadow modeling runs of MST on random $\Dx$ subsets matching the
size of $\Dt$.
Results analysis for these attacks is presented in
section~\ref{subsec:mst-results} of the paper.
We solely note that the $\Dgs$ setting appears to be an in-between of $\Dxs$
and $\Dgh$, in the sense that AUROC and balanced accuracy with simple
activation are similar on $\Dgs$ as on $\Dxs$ save for having higher
standard deviations, while balanced accuracy with calibrated activation on
$\Dgs$ is close to that on $\Dgh$.
This shows that $\Dgs$ samples hardly constitute an easier attack target than
$\Dxs$ ones, so that metrics on these settings are not informative of
worst-case privacy risks. It also hints that calibrated activation is mostly
suited to balanced datasets. Indeed, $\Dgs$ is not perfectly balanced (contrary
to $\Dxs$) due to households grouping a varying number of samples, but is way
closer to being balanced than $\Dxs$.

Three more attacks are added to the comparison.
TAMIS-MST-avg is a variant of TAMIS-MST where scores are computed as the
average, rather than product, of node- and edge-wise terms.
Marginals-$\Sigma$ is a baseline that does not attempt to recover information
about the generative graph. Its attack scores are the average of all possible
1- and 2-way marginal ratio values.
Marginals-$\Pi$ is similar to the previous baseline, but its attack scores
are the product rather than average of 1- and 2-way marginal ratios.

Their respective formulas are
\begin{equation*}
  \begin{aligned}
  \Lambda&_{\textnormal{TAMIS-MST-avg}}(x;G) =\\
    &\frac{1}{|V| + |E|}
    \left(
      \sum_{i \in V}
        \frac{\mu_i^{\Ds}(x)}{\mu_i^{\Dx}(x)}
    + \sum_{(i, j) \in E}
        \frac{\mu_{ij}^{\Ds}(x)}{\mu_{ij}^{\Dx}(x)}
        \frac{\mu_i^{\Dx}(x)\mu_j^{\Dx}(x)}{\mu_i^{\Ds}(x)\mu_j^{\Ds}(x)}
    \right)
  \end{aligned}
\end{equation*}
\begin{equation*}
  \begin{aligned}
    &\Lambda_{\textnormal{Marginals}-\Sigma}(x) =\\
    &\frac{1}{d + d(d - 1)/2}
    \left(
      \sum_{i=1}^d
        \frac{\mu_i^{\Ds}(x)}{\mu_i^{\Dx}(x)}
    + \sum_{i=1}^{d-1} \sum_{j=i+1}^{d}
        \frac{\mu_{ij}^{\Ds}(x)}{\mu_{ij}^{\Dx}(x)}
        \frac{\mu_i^{\Dx}(x)\mu_j^{\Dx}(x)}{\mu_i^{\Ds}(x)\mu_j^{\Ds}(x)}
    \right)
  \end{aligned}
\end{equation*}
\begin{equation*}
    \Lambda_{\textnormal{Marginals}-\Pi}(x) =
    \frac{1}{d + d(d - 1)/2}
    \prod_{i=1}^d
      \left(\frac{\mu_i^{\Ds}(x)}{\mu_i^{\Dx}(x)}\right)^{2 - d}
    \prod_{i=1}^{d-1} \prod_{j=i+1}^{d}
      \frac{\mu_{ij}^{\Ds}(x)}{\mu_{ij}^{\Dx}(x)}
\end{equation*}

Results show that the Marginals-$\Sigma$ and Marginals-$\Pi$ baselines achieve
a lower AUROC and balanced accuracy than the methods compared in main paper,
which once again highlights the value in focusing on distributional features
that were actually modeled by the SDG method rather than on more varied
statistics. We observe that Marginals-$\Sigma$ generally achieves a higher
AUROC and lower balanced accuracy than Marginals-$\Pi$, which is similar to
what we observed in comparing TAMIS-MST with Hybrid-MST and MAMAMIA-MST, that
also differ on whether scores combine information by product or summation.

As for TAMIS-MST-avg, we first observe that it achieves better or similar
metrics as Hybrid-MST, that also uses an averaging of terms but omits 1-way
marginals. We therefore conclude that incorporating 1-way marginals into attack
scores against MST can be beneficial to the attack.
Next, we observe that TAMIS-MST-avg achieves similar or slightly better AUROC
than TAMIS-MST, similar of better balanced accuracy with calibrated activation,
but low balanced accuracy with simple activation. We therefore renew the
conclusion that averaging score terms rather than using a ratio of proper
densities results in scores that have similar theoretical performance, meaning
there exists a decision threshold at which they achieve similar accuracy, but
are not as nicely behaved when using a classic sigmoid activation with default
threshold, making them less suitable for real-life attacks.

\raggedbottom

\begin{table}[H]
  \centering
  \caption{Success of targeted attacks against MST}
  \label{table:results-mst-full}
  \noindent\adjustbox{max width=\textwidth}{%
  \begin{tabular}{llccccc}
  \toprule
   &  & \multicolumn{5}{c}{AUROC} \\
   &  & $\epsilon$=1000 & $\epsilon$=100 & $\epsilon$=10 & $\epsilon$=1 & $\epsilon$=0.1 \\
  \midrule
  \multirow[c]{6}{*}{\Dxs} & TAMIS-MST & \textbf{66.25}\ ($\pm$ 0.4) & 65.53\ ($\pm$ 0.4) & 59.25\ ($\pm$ 0.4) & \textbf{52.57}\ ($\pm$ 0.3) & 50.41\ ($\pm$ 0.3) \\
   & TAMIS-MST-avg & 66.18\ ($\pm$ 0.3) & \textbf{65.62}\ ($\pm$ 0.3) & \textbf{59.91}\ ($\pm$ 0.3) & 52.46\ ($\pm$ 0.3) & 50.36\ ($\pm$ 0.3) \\
   & MAMAMIA-MST & 64.60\ ($\pm$ 1.0) & 64.13\ ($\pm$ 1.0) & 59.05\ ($\pm$ 0.6) & 51.81\ ($\pm$ 0.3) & 50.18\ ($\pm$ 0.3) \\
   & Hybrid-MST & 65.46\ ($\pm$ 0.3) & 64.97\ ($\pm$ 0.3) & 59.59\ ($\pm$ 0.3) & 52.42\ ($\pm$ 0.3) & 50.29\ ($\pm$ 0.3) \\
   & Marginals-$\Sigma$ & 57.21\ ($\pm$ 0.3) & 56.98\ ($\pm$ 0.3) & 54.08\ ($\pm$ 0.3) & 50.59\ ($\pm$ 0.3) & 50.23\ ($\pm$ 0.3) \\
   & Marginals-$\Pi$ & 57.05\ ($\pm$ 0.3) & 57.08\ ($\pm$ 0.3) & 56.40\ ($\pm$ 0.3) & 51.92\ ($\pm$ 0.3) & \textbf{50.61}\ ($\pm$ 0.3) \\
  \cline{1-7}
  \multirow[c]{6}{*}{\Dgs} & TAMIS-MST & 66.82\ ($\pm$ 2.7) & 66.32\ ($\pm$ 2.4) & 60.38\ ($\pm$ 2.7) & 51.63\ ($\pm$ 4.0) & 49.74\ ($\pm$ 3.9) \\
   & TAMIS-MST-avg & \textbf{66.85}\ ($\pm$ 2.8) & \textbf{66.42}\ ($\pm$ 2.7) & \textbf{61.79}\ ($\pm$ 2.8) & 52.44\ ($\pm$ 3.5) & 50.29\ ($\pm$ 2.9) \\
   & MAMAMIA-MST & 65.19\ ($\pm$ 3.0) & 64.92\ ($\pm$ 2.7) & 60.87\ ($\pm$ 3.0) & 51.89\ ($\pm$ 4.0) & 51.03\ ($\pm$ 2.9) \\
   & Hybrid-MST & 66.14\ ($\pm$ 2.8) & 65.81\ ($\pm$ 2.8) & 61.38\ ($\pm$ 3.2) & \textbf{52.54}\ ($\pm$ 4.2) & 50.72\ ($\pm$ 2.7) \\
   & Marginals-$\Sigma$ & 58.75\ ($\pm$ 3.9) & 58.86\ ($\pm$ 3.2) & 56.25\ ($\pm$ 3.7) & 50.47\ ($\pm$ 4.6) & 50.65\ ($\pm$ 4.3) \\
   & Marginals-$\Pi$ & 57.64\ ($\pm$ 4.3) & 57.38\ ($\pm$ 3.9) & 56.96\ ($\pm$ 3.4) & 51.01\ ($\pm$ 4.3) & \textbf{51.06}\ ($\pm$ 4.1) \\
  \cline{1-7}
  \multirow[c]{6}{*}{\Dgh} & TAMIS-MST & \textbf{77.76}\ ($\pm$ 4.8) & \textbf{77.44}\ ($\pm$ 3.6) & 69.87\ ($\pm$ 5.0) & 51.95\ ($\pm$ 6.7) & 49.71\ ($\pm$ 6.1) \\
   & TAMIS-MST-avg & 77.53\ ($\pm$ 5.2) & 77.01\ ($\pm$ 4.4) & \textbf{71.11}\ ($\pm$ 5.0) & \textbf{54.56}\ ($\pm$ 5.0) & 50.60\ ($\pm$ 5.1) \\
   & MAMAMIA-MST & 74.74\ ($\pm$ 5.4) & 74.62\ ($\pm$ 4.5) & 68.51\ ($\pm$ 5.4) & 53.32\ ($\pm$ 6.2) & \textbf{52.09}\ ($\pm$ 5.6) \\
   & Hybrid-MST & 75.97\ ($\pm$ 4.6) & 75.57\ ($\pm$ 4.3) & 69.02\ ($\pm$ 5.5) & 54.46\ ($\pm$ 5.9) & 51.78\ ($\pm$ 5.4) \\
   & Marginals-$\Sigma$ & 63.06\ ($\pm$ 5.2) & 63.10\ ($\pm$ 4.4) & 59.37\ ($\pm$ 5.5) & 51.10\ ($\pm$ 6.7) & 51.38\ ($\pm$ 6.5) \\
   & Marginals-$\Pi$ & 59.69\ ($\pm$ 4.6) & 59.69\ ($\pm$ 4.8) & 60.08\ ($\pm$ 4.9) & 50.72\ ($\pm$ 6.9) & 51.36\ ($\pm$ 5.5) \\
  \bottomrule
  \end{tabular}
  }
  \noindent\adjustbox{max width=\textwidth}{%
  \begin{tabular}{llccccc}
  \toprule
   &  & \multicolumn{5}{c}{Balanced Accuracy (Simple)} \\
   &  & $\epsilon$=1000 & $\epsilon$=100 & $\epsilon$=10 & $\epsilon$=1 & $\epsilon$=0.1 \\
  \midrule
  \multirow[c]{6}{*}{\Dxs} & TAMIS-MST & \textbf{60.84}\ ($\pm$ 0.3) & \textbf{60.39}\ ($\pm$ 0.3) & \textbf{55.66}\ ($\pm$ 0.3) & 50.56\ ($\pm$ 0.1) & 50.01\ ($\pm$ 0.1) \\
   & TAMIS-MST-avg & 53.23\ ($\pm$ 0.2) & 53.14\ ($\pm$ 0.2) & 52.79\ ($\pm$ 0.2) & 51.36\ ($\pm$ 0.2) & \textbf{50.21}\ ($\pm$ 0.3) \\
   & MAMAMIA-MST & 56.66\ ($\pm$ 0.7) & 56.44\ ($\pm$ 0.7) & 54.40\ ($\pm$ 0.4) & 51.15\ ($\pm$ 0.3) & 50.09\ ($\pm$ 0.2) \\
   & Hybrid-MST & 57.27\ ($\pm$ 0.4) & 57.06\ ($\pm$ 0.3) & 54.74\ ($\pm$ 0.3) & \textbf{51.57}\ ($\pm$ 0.3) & 50.19\ ($\pm$ 0.2) \\
   & Marginals-$\Sigma$ & 51.20\ ($\pm$ 0.2) & 51.18\ ($\pm$ 0.1) & 51.40\ ($\pm$ 0.1) & 50.44\ ($\pm$ 0.2) & 50.16\ ($\pm$ 0.3) \\
   & Marginals-$\Pi$ & 53.83\ ($\pm$ 0.3) & 53.84\ ($\pm$ 0.3) & 52.40\ ($\pm$ 0.2) & 50.42\ ($\pm$ 0.1) & 50.03\ ($\pm$ 0.0) \\
  \cline{1-7}
  \multirow[c]{6}{*}{\Dgs} & TAMIS-MST & \textbf{61.62}\ ($\pm$ 2.4) & \textbf{61.54}\ ($\pm$ 2.4) & \textbf{56.94}\ ($\pm$ 1.9) & 50.38\ ($\pm$ 1.1) & 49.97\ ($\pm$ 0.4) \\
   & TAMIS-MST-avg & 52.82\ ($\pm$ 1.3) & 52.76\ ($\pm$ 1.3) & 52.60\ ($\pm$ 1.2) & 51.29\ ($\pm$ 2.2) & 50.19\ ($\pm$ 2.6) \\
   & MAMAMIA-MST & 57.96\ ($\pm$ 2.3) & 57.68\ ($\pm$ 2.2) & 55.21\ ($\pm$ 2.2) & 50.87\ ($\pm$ 2.4) & 50.34\ ($\pm$ 1.6) \\
   & Hybrid-MST & 58.60\ ($\pm$ 2.2) & 58.42\ ($\pm$ 2.4) & 55.42\ ($\pm$ 2.6) & \textbf{51.32}\ ($\pm$ 2.5) & 50.33\ ($\pm$ 1.8) \\
   & Marginals-$\Sigma$ & 51.45\ ($\pm$ 1.2) & 51.22\ ($\pm$ 1.1) & 51.35\ ($\pm$ 1.4) & 50.27\ ($\pm$ 2.6) & \textbf{50.39}\ ($\pm$ 2.7) \\
   & Marginals-$\Pi$ & 54.92\ ($\pm$ 3.1) & 55.34\ ($\pm$ 2.9) & 53.05\ ($\pm$ 1.8) & 50.18\ ($\pm$ 0.9) & 50.04\ ($\pm$ 0.3) \\
  \cline{1-7}
  \multirow[c]{6}{*}{\Dgh} & TAMIS-MST & \textbf{69.86}\ ($\pm$ 4.7) & \textbf{69.52}\ ($\pm$ 4.0) & \textbf{64.26}\ ($\pm$ 4.6) & 50.92\ ($\pm$ 2.9) & 50.02\ ($\pm$ 1.0) \\
   & TAMIS-MST-avg & 52.84\ ($\pm$ 1.8) & 52.56\ ($\pm$ 1.9) & 52.38\ ($\pm$ 1.7) & 52.02\ ($\pm$ 3.8) & 50.72\ ($\pm$ 4.0) \\
   & MAMAMIA-MST & 61.06\ ($\pm$ 4.2) & 60.66\ ($\pm$ 3.4) & 58.16\ ($\pm$ 3.3) & 52.34\ ($\pm$ 4.1) & 50.76\ ($\pm$ 4.0) \\
   & Hybrid-MST & 61.76\ ($\pm$ 3.6) & 61.88\ ($\pm$ 3.7) & 58.08\ ($\pm$ 3.3) & \textbf{52.90}\ ($\pm$ 3.2) & 50.60\ ($\pm$ 4.1) \\
   & Marginals-$\Sigma$ & 51.04\ ($\pm$ 1.4) & 50.86\ ($\pm$ 1.5) & 51.54\ ($\pm$ 1.9) & 51.00\ ($\pm$ 4.0) & \textbf{50.78}\ ($\pm$ 4.5) \\
   & Marginals-$\Pi$ & 55.90\ ($\pm$ 4.3) & 56.08\ ($\pm$ 4.4) & 55.82\ ($\pm$ 4.5) & 50.28\ ($\pm$ 2.7) & 50.10\ ($\pm$ 1.3) \\
  \bottomrule
  \end{tabular}
  }
  \noindent\adjustbox{max width=\textwidth}{%
  \begin{tabular}{llccccc}
  \toprule
   &  & \multicolumn{5}{c}{Balanced Accuracy (Calibrated)} \\
   &  & $\epsilon$=1000 & $\epsilon$=100 & $\epsilon$=10 & $\epsilon$=1 & $\epsilon$=0.1 \\
  \midrule
  \multirow[c]{6}{*}{\Dxs} & TAMIS-MST & \textbf{55.02}\ ($\pm$ 0.2) & \textbf{54.75}\ ($\pm$ 0.2) & 52.63\ ($\pm$ 0.2) & 50.59\ ($\pm$ 0.1) & 50.04\ ($\pm$ 0.1) \\
   & TAMIS-MST-avg & 54.75\ ($\pm$ 0.2) & 54.50\ ($\pm$ 0.2) & 52.64\ ($\pm$ 0.2) & 50.89\ ($\pm$ 0.1) & 50.16\ ($\pm$ 0.1) \\
   & MAMAMIA-MST & 54.38\ ($\pm$ 0.6) & 54.15\ ($\pm$ 0.6) & 52.50\ ($\pm$ 0.4) & 50.89\ ($\pm$ 0.1) & 50.17\ ($\pm$ 0.1) \\
   & Hybrid-MST & 54.76\ ($\pm$ 0.2) & 54.54\ ($\pm$ 0.2) & \textbf{52.69}\ ($\pm$ 0.1) & \textbf{51.08}\ ($\pm$ 0.1) & \textbf{50.27}\ ($\pm$ 0.1) \\
   & Marginals-$\Sigma$ & 51.95\ ($\pm$ 0.2) & 51.86\ ($\pm$ 0.2) & 51.23\ ($\pm$ 0.1) & 50.46\ ($\pm$ 0.1) & 50.17\ ($\pm$ 0.1) \\
   & Marginals-$\Pi$ & 51.43\ ($\pm$ 0.2) & 51.42\ ($\pm$ 0.1) & 51.44\ ($\pm$ 0.1) & 50.55\ ($\pm$ 0.1) & 50.15\ ($\pm$ 0.1) \\
  \cline{1-7}
  \multirow[c]{6}{*}{\Dgs} & TAMIS-MST & 61.75\ ($\pm$ 2.3) & 61.58\ ($\pm$ 2.5) & 57.07\ ($\pm$ 2.2) & 51.26\ ($\pm$ 3.1) & 49.60\ ($\pm$ 3.1) \\
   & TAMIS-MST-avg & \textbf{61.77}\ ($\pm$ 2.2) & \textbf{61.65}\ ($\pm$ 2.3) & \textbf{58.24}\ ($\pm$ 2.1) & 51.64\ ($\pm$ 2.9) & 50.16\ ($\pm$ 2.5) \\
   & MAMAMIA-MST & 60.49\ ($\pm$ 2.3) & 60.17\ ($\pm$ 2.5) & 57.52\ ($\pm$ 2.4) & 51.37\ ($\pm$ 3.4) & 50.69\ ($\pm$ 2.4) \\
   & Hybrid-MST & 61.08\ ($\pm$ 2.3) & 60.80\ ($\pm$ 2.7) & 57.88\ ($\pm$ 2.7) & \textbf{51.86}\ ($\pm$ 3.6) & 50.60\ ($\pm$ 2.2) \\
   & Marginals-$\Sigma$ & 56.23\ ($\pm$ 3.1) & 56.27\ ($\pm$ 2.5) & 54.51\ ($\pm$ 2.9) & 49.99\ ($\pm$ 3.6) & 50.16\ ($\pm$ 3.5) \\
   & Marginals-$\Pi$ & 55.42\ ($\pm$ 3.2) & 55.44\ ($\pm$ 2.9) & 54.70\ ($\pm$ 2.8) & 50.68\ ($\pm$ 3.5) & \textbf{51.10}\ ($\pm$ 3.2) \\
  \cline{1-7}
  \multirow[c]{6}{*}{\Dgh} & TAMIS-MST & \textbf{70.24}\ ($\pm$ 5.2) & 69.88\ ($\pm$ 3.4) & 64.26\ ($\pm$ 4.7) & 51.02\ ($\pm$ 5.3) & 50.08\ ($\pm$ 5.0) \\
   & TAMIS-MST-avg & 70.10\ ($\pm$ 5.1) & \textbf{69.92}\ ($\pm$ 4.5) & \textbf{65.38}\ ($\pm$ 4.8) & 53.10\ ($\pm$ 4.8) & 50.08\ ($\pm$ 4.5) \\
   & MAMAMIA-MST & 67.74\ ($\pm$ 4.8) & 67.24\ ($\pm$ 4.8) & 63.32\ ($\pm$ 5.1) & 52.12\ ($\pm$ 5.4) & \textbf{51.74}\ ($\pm$ 4.5) \\
   & Hybrid-MST & 68.32\ ($\pm$ 4.8) & 68.06\ ($\pm$ 4.2) & 63.56\ ($\pm$ 5.4) & \textbf{53.38}\ ($\pm$ 5.0) & 51.60\ ($\pm$ 4.5) \\
   & Marginals-$\Sigma$ & 58.80\ ($\pm$ 4.6) & 58.72\ ($\pm$ 3.7) & 56.46\ ($\pm$ 4.6) & 50.40\ ($\pm$ 5.6) & 50.80\ ($\pm$ 5.3) \\
   & Marginals-$\Pi$ & 56.68\ ($\pm$ 3.9) & 56.48\ ($\pm$ 4.6) & 58.04\ ($\pm$ 4.0) & 50.66\ ($\pm$ 6.2) & 50.90\ ($\pm$ 4.7) \\
  \bottomrule
  \end{tabular}
  }
\end{table}

\section{Results of attacks against PrivBayes}
\label{appendix:privbayes-results}

This appendix presents exhaustive experimental results about the success of
PrivBayes-targeted attacks, that are reported in
table~\ref{table:results-privbayes-full}.
The evaluated attacks are the same as those presented in the main paper:
TAMIS-PB, Hybrid-PB and MAMAMIA-PB on the one hand, that attempt to recover
the generative bayesian network using either $\Ds$ or $\Dx$; TAMIS-PB* and
Hybrid-PB* on the other hand, that are granted knowledge of the generative
bayesian network.
Reported values are the average and standard deviation of metrics across
50 replicas.
We highlight the highest mean value per setting between attacks in bold,
separating the two groups of attack methods.

Results analysis for these attacks is presented in
section~\ref{subsec:privbayes-results} of the paper.
We solely add comments about results on the $\Dgs$ setting.
First, we observe that the AUROC and balanced accuracy with simple activation
of attacks on $\Dgs$ usually have a slightly higher average value than on
$\Dxs$, and a slightly higher standard deviation. Hence, for PrivBayes, samples
of $\Dg$ appear to be slightly easier to attack than average ones.
Next, we observe that the balanced accuracy with calibrated activation on
$\Dgs$ of MAMAMIA-PB and Hybrid-PB is below that on $\Dgh$ but above that on
$\Dxs$. This renews our conclusion that calibrated activation is mostly
suited to balanced datasets. It also again hints that for PrivBayes $\Dgs$
suffers above-average privacy risks due to correlations unaccounted in DP
guarantees of the SDG process.

\raggedbottom

\begin{table}[H]
  \centering
  \caption{Success of targeted attacks against PrivBayes}
  \label{table:results-privbayes-full}
  \noindent\adjustbox{max width=\textwidth}{%
  \begin{tabular}{llccccc}
  \toprule
   &  & \multicolumn{5}{c}{AUROC} \\
   &  & $\epsilon$=1000 & $\epsilon$=100 & $\epsilon$=10 & $\epsilon$=1 & $\epsilon$=0.1 \\
  \midrule
  \multirow[c]{5}{*}{\Dxs} & TAMIS-PB & 64.65\ ($\pm$ 2.1) & 62.65\ ($\pm$ 1.3) & 53.99\ ($\pm$ 0.3) & 50.72\ ($\pm$ 0.3) & 50.05\ ($\pm$ 0.2) \\
   & MAMAMIA-PB & \textbf{79.34}\ ($\pm$ 1.5) & 64.36\ ($\pm$ 1.1) & \textbf{54.47}\ ($\pm$ 0.3) & \textbf{50.86}\ ($\pm$ 0.3) & 50.12\ ($\pm$ 0.3) \\
   & Hybrid-PB & 79.33\ ($\pm$ 2.7) & \textbf{64.48}\ ($\pm$ 1.7) & 54.42\ ($\pm$ 0.4) & 50.86\ ($\pm$ 0.3) & \textbf{50.12}\ ($\pm$ 0.3) \\
   \cline{2-7}
   & TAMIS-PB* & 66.72\ ($\pm$ 0.8) & 64.61\ ($\pm$ 0.4) & 54.16\ ($\pm$ 0.3) & 50.71\ ($\pm$ 0.3) & 50.05\ ($\pm$ 0.2) \\
   & Hybrid-PB* & \textbf{83.01}\ ($\pm$ 0.9) & \textbf{67.06}\ ($\pm$ 0.3) & \textbf{54.72}\ ($\pm$ 0.2) & \textbf{50.86}\ ($\pm$ 0.3) & \textbf{50.12}\ ($\pm$ 0.3) \\
  \cline{1-7}
  \multirow[c]{5}{*}{\Dgs} & TAMIS-PB & 65.31\ ($\pm$ 3.3) & 64.40\ ($\pm$ 2.8) & 56.71\ ($\pm$ 3.0) & 51.22\ ($\pm$ 3.5) & 50.32\ ($\pm$ 1.8) \\
   & MAMAMIA-PB & \textbf{78.09}\ ($\pm$ 2.5) & 66.27\ ($\pm$ 3.2) & 57.41\ ($\pm$ 2.8) & \textbf{51.56}\ ($\pm$ 3.8) & \textbf{50.78}\ ($\pm$ 4.7) \\
   & Hybrid-PB & 77.91\ ($\pm$ 2.9) & \textbf{66.35}\ ($\pm$ 3.6) & \textbf{57.46}\ ($\pm$ 3.0) & 51.43\ ($\pm$ 3.5) & 50.78\ ($\pm$ 4.7) \\
   \cline{2-7}
   & TAMIS-PB* & 67.57\ ($\pm$ 2.7) & 66.82\ ($\pm$ 2.5) & 56.96\ ($\pm$ 2.8) & 51.22\ ($\pm$ 3.4) & 50.32\ ($\pm$ 1.8) \\
   & Hybrid-PB* & \textbf{81.50}\ ($\pm$ 2.3) & \textbf{69.00}\ ($\pm$ 3.1) & \textbf{57.85}\ ($\pm$ 2.8) & \textbf{51.47}\ ($\pm$ 3.5) & \textbf{50.77}\ ($\pm$ 4.7) \\
  \cline{1-7}
  \multirow[c]{5}{*}{\Dgh} & TAMIS-PB & 82.74\ ($\pm$ 5.6) & 76.85\ ($\pm$ 5.2) & \textbf{62.00}\ ($\pm$ 5.4) & 52.28\ ($\pm$ 6.5) & 50.66\ ($\pm$ 4.2) \\
   & MAMAMIA-PB & \textbf{92.16}\ ($\pm$ 2.6) & 78.06\ ($\pm$ 5.5) & 61.53\ ($\pm$ 4.7) & \textbf{52.44}\ ($\pm$ 6.1) & \textbf{51.30}\ ($\pm$ 6.9) \\
   & Hybrid-PB & 91.87\ ($\pm$ 3.0) & \textbf{78.37}\ ($\pm$ 5.6) & 61.49\ ($\pm$ 5.5) & 52.37\ ($\pm$ 6.2) & 51.30\ ($\pm$ 6.9) \\
   \cline{2-7}
   & TAMIS-PB* & 85.75\ ($\pm$ 4.2) & 80.47\ ($\pm$ 4.8) & \textbf{62.64}\ ($\pm$ 4.7) & \textbf{52.33}\ ($\pm$ 6.5) & 50.66\ ($\pm$ 4.2) \\
   & Hybrid-PB* & \textbf{94.32}\ ($\pm$ 2.6) & \textbf{81.98}\ ($\pm$ 4.8) & 62.23\ ($\pm$ 4.9) & 52.19\ ($\pm$ 6.0) & \textbf{51.31}\ ($\pm$ 6.9) \\
  \bottomrule
  \end{tabular}
  }
  \noindent\adjustbox{max width=\textwidth}{%
  \begin{tabular}{llccccc}
  \toprule
   &  & \multicolumn{5}{c}{Balanced Accuracy (Simple)} \\
   &  & $\epsilon$=1000 & $\epsilon$=100 & $\epsilon$=10 & $\epsilon$=1 & $\epsilon$=0.1 \\
  \midrule
  \multirow[c]{5}{*}{\Dxs} & TAMIS-PB & 61.67\ ($\pm$ 1.8) & \textbf{58.53}\ ($\pm$ 1.1) & \textbf{52.48}\ ($\pm$ 0.3) & \textbf{50.44}\ ($\pm$ 0.2) & 50.02\ ($\pm$ 0.1) \\
   & MAMAMIA-PB & 69.97\ ($\pm$ 1.4) & 55.49\ ($\pm$ 0.6) & 50.81\ ($\pm$ 0.1) & 50.18\ ($\pm$ 0.2) & 50.03\ ($\pm$ 0.2) \\
   & Hybrid-PB & \textbf{70.75}\ ($\pm$ 2.4) & 56.21\ ($\pm$ 0.9) & 50.85\ ($\pm$ 0.1) & 50.21\ ($\pm$ 0.2) & \textbf{50.03}\ ($\pm$ 0.2) \\
   \cline{2-7}
   & TAMIS-PB* & 63.38\ ($\pm$ 0.7) & \textbf{60.18}\ ($\pm$ 0.3) & \textbf{52.68}\ ($\pm$ 0.2) & \textbf{50.45}\ ($\pm$ 0.2) & 50.02\ ($\pm$ 0.1) \\
   & Hybrid-PB* & \textbf{74.24}\ ($\pm$ 1.0) & 57.67\ ($\pm$ 0.4) & 50.90\ ($\pm$ 0.1) & 50.21\ ($\pm$ 0.2) & \textbf{50.03}\ ($\pm$ 0.2) \\
  \cline{1-7}
  \multirow[c]{5}{*}{\Dgs} & TAMIS-PB & 61.99\ ($\pm$ 2.6) & \textbf{60.73}\ ($\pm$ 2.7) & \textbf{54.56}\ ($\pm$ 2.3) & \textbf{50.92}\ ($\pm$ 2.5) & 50.06\ ($\pm$ 0.7) \\
   & MAMAMIA-PB & 68.82\ ($\pm$ 2.5) & 56.94\ ($\pm$ 2.4) & 51.63\ ($\pm$ 1.6) & 50.51\ ($\pm$ 1.5) & 50.30\ ($\pm$ 2.9) \\
   & Hybrid-PB & \textbf{69.70}\ ($\pm$ 3.0) & 58.01\ ($\pm$ 2.5) & 51.72\ ($\pm$ 1.3) & 50.62\ ($\pm$ 1.4) & \textbf{50.30}\ ($\pm$ 3.0) \\
   \cline{2-7}
   & TAMIS-PB* & 63.85\ ($\pm$ 2.5) & \textbf{62.53}\ ($\pm$ 2.4) & \textbf{55.07}\ ($\pm$ 2.0) & \textbf{51.09}\ ($\pm$ 2.3) & 50.06\ ($\pm$ 0.7) \\
   & Hybrid-PB* & \textbf{73.30}\ ($\pm$ 2.1) & 59.66\ ($\pm$ 2.6) & 51.76\ ($\pm$ 1.4) & 50.51\ ($\pm$ 1.5) & \textbf{50.30}\ ($\pm$ 2.9) \\
  \cline{1-7}
  \multirow[c]{5}{*}{\Dgh} & TAMIS-PB & 76.04\ ($\pm$ 5.5) & \textbf{70.20}\ ($\pm$ 4.8) & \textbf{57.04}\ ($\pm$ 4.2) & \textbf{51.72}\ ($\pm$ 5.8) & 50.08\ ($\pm$ 1.8) \\
   & MAMAMIA-PB & 78.54\ ($\pm$ 4.2) & 57.34\ ($\pm$ 3.7) & 51.74\ ($\pm$ 2.5) & 50.46\ ($\pm$ 2.2) & \textbf{50.18}\ ($\pm$ 4.1) \\
   & Hybrid-PB & \textbf{82.14}\ ($\pm$ 4.9) & 59.86\ ($\pm$ 3.6) & 51.70\ ($\pm$ 2.4) & 50.16\ ($\pm$ 2.3) & 50.14\ ($\pm$ 4.2) \\
   \cline{2-7}
   & TAMIS-PB* & 79.48\ ($\pm$ 4.8) & \textbf{73.12}\ ($\pm$ 4.5) & \textbf{57.64}\ ($\pm$ 3.5) & \textbf{51.60}\ ($\pm$ 5.0) & 50.08\ ($\pm$ 1.8) \\
   & Hybrid-PB* & \textbf{86.58}\ ($\pm$ 4.0) & 62.36\ ($\pm$ 3.4) & 51.54\ ($\pm$ 2.3) & 50.26\ ($\pm$ 2.3) & \textbf{50.22}\ ($\pm$ 4.1) \\
  \bottomrule
  \end{tabular}
  }
  \noindent\adjustbox{max width=\textwidth}{%
  \begin{tabular}{llccccc}
  \toprule
   &  & \multicolumn{5}{c}{Balanced Accuracy (Calibrated)} \\
   &  & $\epsilon$=1000 & $\epsilon$=100 & $\epsilon$=10 & $\epsilon$=1 & $\epsilon$=0.1 \\
  \midrule
  \multirow[c]{5}{*}{\Dxs} & TAMIS-PB & 59.88\ ($\pm$ 1.6) & \textbf{54.03}\ ($\pm$ 0.7) & \textbf{50.92}\ ($\pm$ 0.1) & \textbf{50.19}\ ($\pm$ 0.1) & 50.01\ ($\pm$ 0.1) \\
   & MAMAMIA-PB & \textbf{62.02}\ ($\pm$ 1.2) & 53.59\ ($\pm$ 0.4) & 50.68\ ($\pm$ 0.1) & 50.12\ ($\pm$ 0.1) & 50.02\ ($\pm$ 0.1) \\
   & Hybrid-PB & 61.94\ ($\pm$ 1.7) & 53.65\ ($\pm$ 0.6) & 50.67\ ($\pm$ 0.1) & 50.12\ ($\pm$ 0.1) & \textbf{50.02}\ ($\pm$ 0.1) \\
   \cline{2-7}
   & TAMIS-PB* & 61.62\ ($\pm$ 0.5) & \textbf{55.02}\ ($\pm$ 0.2) & \textbf{51.01}\ ($\pm$ 0.1) & \textbf{50.18}\ ($\pm$ 0.1) & 50.01\ ($\pm$ 0.1) \\
   & Hybrid-PB* & \textbf{64.32}\ ($\pm$ 0.7) & 54.66\ ($\pm$ 0.2) & 50.73\ ($\pm$ 0.1) & 50.12\ ($\pm$ 0.1) & \textbf{50.02}\ ($\pm$ 0.1) \\
  \cline{1-7}
  \multirow[c]{5}{*}{\Dgs} & TAMIS-PB & 50.00\ ($\pm$ 0.0) & 60.63\ ($\pm$ 2.6) & 54.97\ ($\pm$ 2.4) & 51.15\ ($\pm$ 2.6) & 50.00\ ($\pm$ 0.0) \\
   & MAMAMIA-PB & \textbf{70.96}\ ($\pm$ 2.3) & 61.31\ ($\pm$ 2.8) & 55.24\ ($\pm$ 2.6) & \textbf{51.30}\ ($\pm$ 3.0) & 50.59\ ($\pm$ 3.7) \\
   & Hybrid-PB & 70.81\ ($\pm$ 2.5) & \textbf{61.61}\ ($\pm$ 3.1) & \textbf{55.34}\ ($\pm$ 2.6) & 51.15\ ($\pm$ 2.9) & \textbf{50.61}\ ($\pm$ 3.7) \\
   \cline{2-7}
   & TAMIS-PB* & 50.00\ ($\pm$ 0.0) & 62.70\ ($\pm$ 2.1) & 55.24\ ($\pm$ 2.3) & 51.16\ ($\pm$ 2.5) & 50.00\ ($\pm$ 0.0) \\
   & Hybrid-PB* & \textbf{73.94}\ ($\pm$ 2.3) & \textbf{63.40}\ ($\pm$ 2.8) & \textbf{55.97}\ ($\pm$ 2.4) & \textbf{51.26}\ ($\pm$ 2.8) & \textbf{50.60}\ ($\pm$ 3.7) \\
  \cline{1-7}
  \multirow[c]{5}{*}{\Dgh} & TAMIS-PB & 76.26\ ($\pm$ 5.5) & 70.18\ ($\pm$ 5.3) & \textbf{58.88}\ ($\pm$ 5.0) & 51.44\ ($\pm$ 4.8) & 50.36\ ($\pm$ 2.6) \\
   & MAMAMIA-PB & \textbf{84.36}\ ($\pm$ 4.2) & 71.04\ ($\pm$ 5.3) & 58.44\ ($\pm$ 4.6) & \textbf{51.96}\ ($\pm$ 5.2) & \textbf{51.10}\ ($\pm$ 5.9) \\
   & Hybrid-PB & 84.14\ ($\pm$ 4.5) & \textbf{71.44}\ ($\pm$ 5.6) & 58.52\ ($\pm$ 5.0) & 51.76\ ($\pm$ 4.8) & 51.10\ ($\pm$ 6.0) \\
   \cline{2-7}
   & TAMIS-PB* & 79.28\ ($\pm$ 4.9) & 73.40\ ($\pm$ 4.6) & \textbf{59.34}\ ($\pm$ 4.2) & 51.36\ ($\pm$ 4.7) & 50.36\ ($\pm$ 2.6) \\
   & Hybrid-PB* & \textbf{86.86}\ ($\pm$ 3.9) & \textbf{74.24}\ ($\pm$ 4.8) & 59.08\ ($\pm$ 4.5) & \textbf{51.84}\ ($\pm$ 4.9) & \textbf{51.14}\ ($\pm$ 6.0) \\
  \bottomrule
  \end{tabular}
  }
\end{table}

\section{Results of attacks using cross-targeted methods}
\label{appendix:crosstarget-results}

This appendix presents exhaustive experimental results about the success of
PrivBayes-targeted attacks against MST-generated data, that are reported in
table~\ref{table:results-cross-mst-full}, and that of MST-targeted attacks
against PrivBayes-generated data, that are reported in
table~\ref{table:results-cross-privbayes-full}.
Reported values are the average and standard deviation of metrics across
50 replicas.
We highlight the highest mean value per setting between attacks in bold.

The aim of these experiments was to assess the extent to which black-box
knowledge of the SDG method is key to achieve high MIA success. The hypothesis
on which our attacks and the MAMA-MIA one are rooted is that tailoring density
estimators to the SDG method is key to achieving high success. However, we also
observed that at the same formal DP level, attacks against MST and PrivBayes
do not achieve the same success. This could be due to one attack class being
better than the other, which we verify empirically with these cross-targeted
experiments.

These experiments amount to an attacker being mistaken about the SDG method
that generated $\Ds$. For TAMIS and Hybrid attacks, we learned a tree graph
from each PrivBayes-generated $\Ds$, and a bayesian network from each
MST-generated $\Ds$ using default PrivBayes hyper-parameters and the
$\epsilon$ value associated with that dataset. For MAMA-MIA attacks, we re-used
shadow modeling weights from our main experiments.

As summarized in section~\ref{subsec:cross-results} of the paper, we mainly
conclude that cross-targeted attacks achieve markedly lower success than
properly-targeted ones.

\subsection*{PrivBayes-targeted attacks against MST}

For PrivBayes-targeted attacks against MST, we observe that attacks are
unsuccessful for low $\epsilon$ values, but also under-perform for
$\epsilon=1000$. While the former is due to DP guarantees, we believe that the
latter is due to the learned bayesian network or shadow modeling weights
covering too many complex dependencies, making attack scores un-tailored.
For $\epsilon \in [100, 10]$, attacks are more successful than the baseline
ones reported in Appendix~\ref{appendix:mst-results}, but less so than their
properly-targeted counterparts. All three attacks have close results, with
TAMIS-PB achieving the best accuracy with simple activation and Hybrid-PB
the best one with calibrated activation.

Overall, we conclude that PrivBayes-targeted attacks generalize poorly to
MST-generated data, due to the added complexity of bayesian networks
resulting in attack scores that are less tailored to the generative tree
graph than MST-targeted attacks.

\begin{table}[H]
  \centering
  \caption{Success of PrivBayes-targeted attacks against MST}
  \label{table:results-cross-mst-full}
  \noindent\adjustbox{max width=\textwidth}{%
  \begin{tabular}{llccccc}
  \toprule
   &  & \multicolumn{5}{c}{AUROC} \\
   &  & $\epsilon$=1000 & $\epsilon$=100 & $\epsilon$=10 & $\epsilon$=1 & $\epsilon$=0.1 \\
  \midrule
  \multirow[c]{3}{*}{\Dxs} & TAMIS-PB & 51.75\ ($\pm$ 0.3) & 58.28\ ($\pm$ 0.4) & \textbf{53.39}\ ($\pm$ 0.3) & \textbf{50.20}\ ($\pm$ 0.3) & \textbf{50.27}\ ($\pm$ 0.3) \\
   & MAMAMIA-PB & \textbf{56.17}\ ($\pm$ 0.4) & 59.32\ ($\pm$ 0.5) & 52.98\ ($\pm$ 0.3) & 50.17\ ($\pm$ 0.3) & 50.15\ ($\pm$ 0.3) \\
   & Hybrid-PB & 55.98\ ($\pm$ 1.1) & \textbf{59.50}\ ($\pm$ 0.4) & 53.32\ ($\pm$ 0.3) & 50.17\ ($\pm$ 0.3) & 50.15\ ($\pm$ 0.3) \\
  \cline{1-7}
  \multirow[c]{3}{*}{\Dgs} & TAMIS-PB & 51.25\ ($\pm$ 3.0) & 57.81\ ($\pm$ 4.2) & 57.15\ ($\pm$ 2.9) & \textbf{50.17}\ ($\pm$ 4.8) & \textbf{50.91}\ ($\pm$ 4.4) \\
   & MAMAMIA-PB & 54.99\ ($\pm$ 3.7) & 59.30\ ($\pm$ 2.7) & 56.37\ ($\pm$ 2.8) & 50.13\ ($\pm$ 4.4) & 50.67\ ($\pm$ 4.6) \\
   & Hybrid-PB & \textbf{55.57}\ ($\pm$ 3.8) & \textbf{59.78}\ ($\pm$ 3.3) & \textbf{57.34}\ ($\pm$ 3.0) & 50.16\ ($\pm$ 4.6) & 50.67\ ($\pm$ 4.6) \\
  \cline{1-7}
  \multirow[c]{3}{*}{\Dgh} & TAMIS-PB & 53.77\ ($\pm$ 6.5) & 65.22\ ($\pm$ 6.8) & 60.78\ ($\pm$ 5.2) & 49.71\ ($\pm$ 6.0) & \textbf{51.80}\ ($\pm$ 6.8) \\
   & MAMAMIA-PB & 58.83\ ($\pm$ 6.4) & 66.22\ ($\pm$ 4.5) & 60.10\ ($\pm$ 4.9) & 50.34\ ($\pm$ 6.3) & 51.27\ ($\pm$ 6.6) \\
   & Hybrid-PB & \textbf{59.64}\ ($\pm$ 6.7) & \textbf{67.88}\ ($\pm$ 5.9) & \textbf{60.99}\ ($\pm$ 5.4) & \textbf{50.40}\ ($\pm$ 6.5) & 51.26\ ($\pm$ 6.6) \\
  \bottomrule
  \end{tabular}
  }
  \noindent\adjustbox{max width=\textwidth}{%
  \begin{tabular}{llccccc}
  \toprule
   &  & \multicolumn{5}{c}{Balanced Accuracy (Simple)} \\
   &  & $\epsilon$=1000 & $\epsilon$=100 & $\epsilon$=10 & $\epsilon$=1 & $\epsilon$=0.1 \\
  \midrule
  \multirow[c]{3}{*}{\Dxs} & TAMIS-PB & 51.10\ ($\pm$ 0.2) & \textbf{55.35}\ ($\pm$ 0.3) & \textbf{52.01}\ ($\pm$ 0.2) & 50.10\ ($\pm$ 0.2) & \textbf{50.09}\ ($\pm$ 0.1) \\
   & MAMAMIA-PB & 52.43\ ($\pm$ 0.3) & 53.71\ ($\pm$ 0.3) & 50.82\ ($\pm$ 0.2) & \textbf{50.17}\ ($\pm$ 0.2) & 50.09\ ($\pm$ 0.2) \\
   & Hybrid-PB & \textbf{52.60}\ ($\pm$ 0.5) & 54.04\ ($\pm$ 0.2) & 50.90\ ($\pm$ 0.1) & 50.17\ ($\pm$ 0.2) & 50.09\ ($\pm$ 0.2) \\
  \cline{1-7}
  \multirow[c]{3}{*}{\Dgs} & TAMIS-PB & 51.01\ ($\pm$ 1.6) & \textbf{55.61}\ ($\pm$ 2.8) & \textbf{54.52}\ ($\pm$ 2.4) & 49.95\ ($\pm$ 2.3) & 50.44\ ($\pm$ 1.3) \\
   & MAMAMIA-PB & 51.33\ ($\pm$ 1.8) & 53.63\ ($\pm$ 1.5) & 51.79\ ($\pm$ 1.8) & \textbf{50.32}\ ($\pm$ 2.6) & \textbf{50.48}\ ($\pm$ 2.9) \\
   & Hybrid-PB & \textbf{52.17}\ ($\pm$ 2.1) & 54.46\ ($\pm$ 2.2) & 51.84\ ($\pm$ 1.8) & 50.30\ ($\pm$ 2.7) & 50.44\ ($\pm$ 2.9) \\
  \cline{1-7}
  \multirow[c]{3}{*}{\Dgh} & TAMIS-PB & 51.74\ ($\pm$ 3.4) & \textbf{60.26}\ ($\pm$ 5.1) & \textbf{56.04}\ ($\pm$ 4.5) & 50.04\ ($\pm$ 4.1) & 50.42\ ($\pm$ 2.7) \\
   & MAMAMIA-PB & 51.72\ ($\pm$ 2.6) & 54.12\ ($\pm$ 2.6) & 51.50\ ($\pm$ 2.3) & 50.30\ ($\pm$ 3.0) & 50.88\ ($\pm$ 3.8) \\
   & Hybrid-PB & \textbf{52.34}\ ($\pm$ 3.1) & 55.48\ ($\pm$ 3.2) & 51.72\ ($\pm$ 2.3) & \textbf{50.44}\ ($\pm$ 3.4) & \textbf{50.90}\ ($\pm$ 3.9) \\
  \bottomrule
  \end{tabular}
  }
  \noindent\adjustbox{max width=\textwidth}{%
  \begin{tabular}{llccccc}
  \toprule
   &  & \multicolumn{5}{c}{Balanced Accuracy (Calibrated)} \\
   &  & $\epsilon$=1000 & $\epsilon$=100 & $\epsilon$=10 & $\epsilon$=1 & $\epsilon$=0.1 \\
  \midrule
  \multirow[c]{3}{*}{\Dxs} & TAMIS-PB & 50.83\ ($\pm$ 0.2) & 52.13\ ($\pm$ 0.2) & \textbf{50.76}\ ($\pm$ 0.1) & 50.11\ ($\pm$ 0.1) & \textbf{50.08}\ ($\pm$ 0.1) \\
   & MAMAMIA-PB & 50.90\ ($\pm$ 0.2) & 52.02\ ($\pm$ 0.2) & 50.58\ ($\pm$ 0.1) & \textbf{50.15}\ ($\pm$ 0.1) & 50.06\ ($\pm$ 0.1) \\
   & Hybrid-PB & \textbf{50.98}\ ($\pm$ 0.3) & \textbf{52.13}\ ($\pm$ 0.2) & 50.71\ ($\pm$ 0.1) & 50.15\ ($\pm$ 0.1) & 50.06\ ($\pm$ 0.1) \\
  \cline{1-7}
  \multirow[c]{3}{*}{\Dgs} & TAMIS-PB & 50.00\ ($\pm$ 0.0) & 55.78\ ($\pm$ 3.3) & 55.10\ ($\pm$ 2.3) & \textbf{49.89}\ ($\pm$ 3.8) & \textbf{50.60}\ ($\pm$ 3.4) \\
   & MAMAMIA-PB & 53.55\ ($\pm$ 2.9) & 56.33\ ($\pm$ 2.3) & 54.45\ ($\pm$ 2.4) & 49.84\ ($\pm$ 3.6) & 50.14\ ($\pm$ 3.5) \\
   & Hybrid-PB & \textbf{53.75}\ ($\pm$ 2.9) & \textbf{56.88}\ ($\pm$ 2.6) & \textbf{55.21}\ ($\pm$ 2.6) & 49.79\ ($\pm$ 3.7) & 50.12\ ($\pm$ 3.5) \\
  \cline{1-7}
  \multirow[c]{3}{*}{\Dgh} & TAMIS-PB & 53.36\ ($\pm$ 4.9) & 60.90\ ($\pm$ 5.8) & 57.44\ ($\pm$ 4.4) & 49.32\ ($\pm$ 4.6) & 50.90\ ($\pm$ 5.7) \\
   & MAMAMIA-PB & 55.90\ ($\pm$ 5.4) & 61.50\ ($\pm$ 4.3) & 57.12\ ($\pm$ 4.6) & 50.16\ ($\pm$ 5.6) & \textbf{50.92}\ ($\pm$ 4.9) \\
   & Hybrid-PB & \textbf{56.36}\ ($\pm$ 5.5) & \textbf{62.30}\ ($\pm$ 5.4) & \textbf{57.68}\ ($\pm$ 4.5) & \textbf{50.20}\ ($\pm$ 5.7) & 50.70\ ($\pm$ 4.9) \\
  \bottomrule
  \end{tabular}
  }
\end{table}

\subsection*{MST-targeted attacks against PrivBayes}

For MST-targeted attacks against PrivBayes, we observe that attacks are
unsuccessful for low $\epsilon$ values due to DP guarantees, but that some
achieve significative success in higher-$\epsilon$ settings.

For $\epsilon \in [100, 10]$, TAMIS-MST achieves similar or better balanced
accuracy than MAMAMIA-PB did (cf. values reported in
Appendix~\ref{appendix:privbayes-results}), albeit with increased standard
deviation. Its AUROC is smaller than that of PrivBayes-targeted attacks,
but only by 2 to 4 percentage points, which is significative but still means
TAMIS-MST is not such a bad attack in these settings.
However, for $\epsilon=1000$, TAMIS-MST does not rival with the exceptional
performance of some of its targeted counterparts, which we hypothesize to be
due to the exceedingly simple nature of tree graphical models compared with
bayesian networks in such a setting where parent sets are allowed to be wide.

Hybrid-MST and MAMAMIA-MST achieve close but lower performance than TAMIS-MST
against PrivBayes-generated data. As for the TAMIS-MST-avg variant, it is close
to TAMIS-MST but fails to achieve significative balanced accuracy with simple
activation.

Overall, we conclude that MST-targeted attacks can partially generalize to
PrivBayes-generated data, albeit with worse results than their
properly-targeted counterparts. We hypothesize that this is due to the tree
graphical model covering part of, but not all, information that was used by
the SDG method to generate $\Ds$, resulting in attack scores being based on
relevant but un-exhaustive statistics.

We note that TAMIS-MST was the most competitive cross-targeted attack, which
is interesting due to it not requiring any knowledge of the SDG method,
contrary to MAMAMIA-MST using knowledge of the privacy budget for shadow
modeling. Indeed, this characteristic means that TAMIS-MST could realistically
be applied in no-box model threats where the attacker is agnostic to the SDG
method. Further work would however be required to assess whether TAMIS-MST
would constitute a good attack in such a setting, by comparing it with
relevant baselines and considering more various SDG methods to attack.

\begin{table}[H]
  \centering
  \caption{Success of MST-targeted attacks against PrivBayes}
  \label{table:results-cross-privbayes-full}
  \noindent\adjustbox{max width=\textwidth}{%
  \begin{tabular}{llccccc}
  \toprule
   &  & \multicolumn{5}{c}{AUROC} \\
   &  & $\epsilon$=1000 & $\epsilon$=100 & $\epsilon$=10 & $\epsilon$=1 & $\epsilon$=0.1 \\
  \midrule
  \multirow[c]{4}{*}{\Dxs} & TAMIS-MST & \textbf{62.07}\ ($\pm$ 1.2) & 60.03\ ($\pm$ 1.3) & 52.06\ ($\pm$ 0.4) & 50.55\ ($\pm$ 0.3) & 49.99\ ($\pm$ 0.3) \\
   & TAMIS-MST-avg & 61.54\ ($\pm$ 0.7) & \textbf{60.19}\ ($\pm$ 1.2) & 52.11\ ($\pm$ 0.4) & 50.34\ ($\pm$ 0.3) & 49.84\ ($\pm$ 0.3) \\
   & MAMAMIA-MST & 59.41\ ($\pm$ 0.5) & 59.62\ ($\pm$ 1.1) & 52.42\ ($\pm$ 0.3) & 50.53\ ($\pm$ 0.3) & \textbf{50.09}\ ($\pm$ 0.3) \\
   & Hybrid-MST & 60.97\ ($\pm$ 0.7) & 59.91\ ($\pm$ 1.1) & \textbf{52.55}\ ($\pm$ 0.4) & \textbf{50.64}\ ($\pm$ 0.3) & 50.07\ ($\pm$ 0.3) \\
  \cline{1-7}
  \multirow[c]{4}{*}{\Dgs} & TAMIS-MST & \textbf{64.16}\ ($\pm$ 2.9) & 61.91\ ($\pm$ 3.4) & 54.46\ ($\pm$ 3.6) & 50.81\ ($\pm$ 3.4) & \textbf{50.89}\ ($\pm$ 3.9) \\
   & TAMIS-MST-avg & 64.02\ ($\pm$ 3.1) & \textbf{62.17}\ ($\pm$ 3.7) & 54.69\ ($\pm$ 3.3) & 51.06\ ($\pm$ 3.8) & 49.62\ ($\pm$ 3.3) \\
   & MAMAMIA-MST & 61.12\ ($\pm$ 3.6) & 61.63\ ($\pm$ 3.3) & 55.15\ ($\pm$ 3.5) & 51.34\ ($\pm$ 4.3) & 50.66\ ($\pm$ 3.7) \\
   & Hybrid-MST & 63.53\ ($\pm$ 3.4) & 61.91\ ($\pm$ 3.3) & \textbf{55.26}\ ($\pm$ 3.2) & \textbf{51.43}\ ($\pm$ 3.7) & 50.62\ ($\pm$ 3.8) \\
  \cline{1-7}
  \multirow[c]{4}{*}{\Dgh} & TAMIS-MST & \textbf{72.58}\ ($\pm$ 5.3) & \textbf{70.60}\ ($\pm$ 5.9) & \textbf{57.46}\ ($\pm$ 6.5) & 51.74\ ($\pm$ 5.6) & 50.95\ ($\pm$ 5.4) \\
   & TAMIS-MST-avg & 71.73\ ($\pm$ 4.6) & 70.40\ ($\pm$ 6.6) & 56.04\ ($\pm$ 6.1) & 49.87\ ($\pm$ 6.3) & 49.20\ ($\pm$ 5.6) \\
   & MAMAMIA-MST & 66.80\ ($\pm$ 5.7) & 68.83\ ($\pm$ 5.1) & 57.02\ ($\pm$ 6.0) & 51.94\ ($\pm$ 7.4) & 51.75\ ($\pm$ 5.9) \\
   & Hybrid-MST & 70.41\ ($\pm$ 5.2) & 69.45\ ($\pm$ 5.5) & 57.21\ ($\pm$ 6.0) & \textbf{52.51}\ ($\pm$ 7.2) & \textbf{51.84}\ ($\pm$ 6.4) \\
  \bottomrule
  \end{tabular}
  }
  \noindent\adjustbox{max width=\textwidth}{%
  \begin{tabular}{llccccc}
  \toprule
   &  & \multicolumn{5}{c}{Balanced Accuracy (Simple)} \\
   &  & $\epsilon$=1000 & $\epsilon$=100 & $\epsilon$=10 & $\epsilon$=1 & $\epsilon$=0.1 \\
  \midrule
  \multirow[c]{4}{*}{\Dxs} & TAMIS-MST & \textbf{57.29}\ ($\pm$ 0.8) & \textbf{56.35}\ ($\pm$ 0.9) & \textbf{51.09}\ ($\pm$ 0.3) & 50.26\ ($\pm$ 0.2) & 49.97\ ($\pm$ 0.2) \\
   & TAMIS-MST-avg & 51.75\ ($\pm$ 0.2) & 51.23\ ($\pm$ 0.2) & 50.18\ ($\pm$ 0.1) & 49.97\ ($\pm$ 0.2) & 49.95\ ($\pm$ 0.2) \\
   & MAMAMIA-MST & 53.15\ ($\pm$ 0.4) & 53.52\ ($\pm$ 0.5) & 50.69\ ($\pm$ 0.2) & 50.18\ ($\pm$ 0.2) & \textbf{50.01}\ ($\pm$ 0.1) \\
   & Hybrid-MST & 54.68\ ($\pm$ 0.4) & 53.62\ ($\pm$ 0.5) & 50.83\ ($\pm$ 0.2) & \textbf{50.33}\ ($\pm$ 0.3) & 50.00\ ($\pm$ 0.1) \\
  \cline{1-7}
  \multirow[c]{4}{*}{\Dgs} & TAMIS-MST & \textbf{59.58}\ ($\pm$ 2.3) & \textbf{58.27}\ ($\pm$ 2.5) & \textbf{52.73}\ ($\pm$ 2.4) & \textbf{50.88}\ ($\pm$ 2.2) & 50.15\ ($\pm$ 1.3) \\
   & TAMIS-MST-avg & 52.04\ ($\pm$ 1.5) & 51.44\ ($\pm$ 1.4) & 50.31\ ($\pm$ 1.1) & 50.13\ ($\pm$ 2.6) & 50.12\ ($\pm$ 2.6) \\
   & MAMAMIA-MST & 54.41\ ($\pm$ 2.4) & 55.20\ ($\pm$ 2.3) & 51.64\ ($\pm$ 1.9) & 50.66\ ($\pm$ 2.2) & 50.23\ ($\pm$ 1.4) \\
   & Hybrid-MST & 56.88\ ($\pm$ 2.6) & 55.36\ ($\pm$ 2.4) & 51.96\ ($\pm$ 2.1) & 50.73\ ($\pm$ 2.6) & \textbf{50.39}\ ($\pm$ 1.5) \\
  \cline{1-7}
  \multirow[c]{4}{*}{\Dgh} & TAMIS-MST & \textbf{65.76}\ ($\pm$ 4.6) & \textbf{64.78}\ ($\pm$ 5.6) & \textbf{53.96}\ ($\pm$ 4.4) & 51.28\ ($\pm$ 4.0) & 50.38\ ($\pm$ 4.0) \\
   & TAMIS-MST-avg & 51.46\ ($\pm$ 1.9) & 51.00\ ($\pm$ 1.5) & 50.06\ ($\pm$ 1.8) & 48.80\ ($\pm$ 4.3) & 50.16\ ($\pm$ 1.9) \\
   & MAMAMIA-MST & 54.86\ ($\pm$ 2.9) & 56.00\ ($\pm$ 3.8) & 51.82\ ($\pm$ 2.9) & 50.80\ ($\pm$ 3.5) & 50.62\ ($\pm$ 3.1) \\
   & Hybrid-MST & 57.96\ ($\pm$ 3.7) & 56.02\ ($\pm$ 3.5) & 51.78\ ($\pm$ 3.5) & \textbf{51.46}\ ($\pm$ 4.5) & \textbf{50.80}\ ($\pm$ 3.1) \\
  \bottomrule
  \end{tabular}
  }
  \noindent\adjustbox{max width=\textwidth}{%
  \begin{tabular}{llccccc}
  \toprule
   &  & \multicolumn{5}{c}{Balanced Accuracy (Calibrated)} \\
   &  & $\epsilon$=1000 & $\epsilon$=100 & $\epsilon$=10 & $\epsilon$=1 & $\epsilon$=0.1 \\
  \midrule
  \multirow[c]{4}{*}{\Dxs} & TAMIS-MST & \textbf{53.02}\ ($\pm$ 0.4) & \textbf{52.42}\ ($\pm$ 0.4) & \textbf{50.44}\ ($\pm$ 0.1) & 50.09\ ($\pm$ 0.1) & 49.97\ ($\pm$ 0.1) \\
   & TAMIS-MST-avg & 52.38\ ($\pm$ 0.3) & 51.95\ ($\pm$ 0.3) & 50.18\ ($\pm$ 0.1) & 49.93\ ($\pm$ 0.1) & 49.97\ ($\pm$ 0.1) \\
   & MAMAMIA-MST & 51.92\ ($\pm$ 0.2) & 51.93\ ($\pm$ 0.3) & 50.34\ ($\pm$ 0.1) & 50.07\ ($\pm$ 0.1) & 50.01\ ($\pm$ 0.1) \\
   & Hybrid-MST & 52.41\ ($\pm$ 0.3) & 52.00\ ($\pm$ 0.3) & 50.34\ ($\pm$ 0.1) & \textbf{50.11}\ ($\pm$ 0.1) & \textbf{50.01}\ ($\pm$ 0.1) \\
  \cline{1-7}
  \multirow[c]{4}{*}{\Dgs} & TAMIS-MST & \textbf{60.11}\ ($\pm$ 2.3) & 58.60\ ($\pm$ 2.8) & 53.19\ ($\pm$ 2.7) & 50.44\ ($\pm$ 2.7) & 50.30\ ($\pm$ 3.0) \\
   & TAMIS-MST-avg & 59.82\ ($\pm$ 2.5) & \textbf{58.74}\ ($\pm$ 3.0) & 53.68\ ($\pm$ 2.6) & 50.91\ ($\pm$ 3.2) & 49.84\ ($\pm$ 2.8) \\
   & MAMAMIA-MST & 57.76\ ($\pm$ 3.0) & 58.03\ ($\pm$ 2.5) & \textbf{53.91}\ ($\pm$ 2.6) & 50.88\ ($\pm$ 3.3) & \textbf{50.47}\ ($\pm$ 3.1) \\
   & Hybrid-MST & 59.44\ ($\pm$ 2.5) & 58.31\ ($\pm$ 2.8) & 53.76\ ($\pm$ 2.6) & \textbf{51.15}\ ($\pm$ 3.0) & 50.32\ ($\pm$ 3.2) \\
  \cline{1-7}
  \multirow[c]{4}{*}{\Dgh} & TAMIS-MST & \textbf{65.96}\ ($\pm$ 4.9) & 64.40\ ($\pm$ 5.5) & \textbf{55.56}\ ($\pm$ 5.3) & 50.82\ ($\pm$ 4.5) & \textbf{51.36}\ ($\pm$ 4.7) \\
   & TAMIS-MST-avg & 65.40\ ($\pm$ 4.2) & \textbf{64.56}\ ($\pm$ 6.2) & 55.00\ ($\pm$ 5.4) & 49.90\ ($\pm$ 5.7) & 49.68\ ($\pm$ 5.4) \\
   & MAMAMIA-MST & 61.72\ ($\pm$ 4.5) & 63.64\ ($\pm$ 5.0) & 54.92\ ($\pm$ 5.0) & \textbf{51.48}\ ($\pm$ 5.7) & 50.96\ ($\pm$ 4.7) \\
   & Hybrid-MST & 64.64\ ($\pm$ 4.4) & 63.68\ ($\pm$ 4.7) & 55.36\ ($\pm$ 5.1) & 51.36\ ($\pm$ 5.5) & 51.00\ ($\pm$ 5.3) \\
  \bottomrule
  \end{tabular}
  }
\end{table}

\end{document}